\pdfoutput=1
\documentclass[authoryear, preprint, 12pt]{elsarticle}

\usepackage{graphicx}
\usepackage{amsmath,amssymb,amsthm}
\usepackage{hyperref}

\usepackage{tikz}
\usetikzlibrary{arrows,arrows.meta,backgrounds,calc,patterns,positioning,shapes,shadows}
\usepackage[linesnumbered, ruled, noend]{algorithm2e}
\usepackage[capitalise]{cleveref}

\crefname{algorithm}{Alg.}{Algs.}
\Crefname{algorithm}{Algorithm}{Algorithms}

\newtheorem{definition}{Definition}[section]
\newtheorem{example}{Example}[section]
\newtheorem{theorem}{Theorem}[section]

\usepackage{csquotes}
\usepackage{subcaption}
\usepackage[nolist, nohyperlinks]{acronym}
\acrodef{acp}[ACP]{advanced colour passing}
\acrodef{cp}[CP]{colour passing}
\acrodef{fg}[FG]{factor graph}
\acrodef{kl}[KL]{Kullback-Leibler}
\acrodef{kld}[KLD]{Kullback-Leibler divergence}
\acrodef{lifg}[LIFAGU]{Lifting Factor Graphs with Some Unknown Factors}
\acrodef{ljt}[LJT]{lifted junction tree}
\acrodef{lv}[logvar]{logical variable}
\acrodef{lve}[LVE]{lifted variable elimination}
\acrodef{pf}[parfactor]{parametric factor}
\acrodef{pfg}[PFG]{parameterised factor graph}
\acrodef{prv}[PRV]{parameterised randvar}
\acrodef{rv}[randvar]{random variable}
\acrodef{ve}[VE]{variable elimination}
\acrodef{wl}[WL]{Weisfeiler-Leman}

\newcommand{\abs}[1]{\lvert #1 \rvert}
\newcommand{\domain}[1]{\ensuremath{\mathrm{dom}(#1)}}
\newcommand{\KLD}{\ensuremath{\mathrm{KLD}}}
\newcommand{\twostep}[1]{\ensuremath{2\text{-}step(#1)}}
\newcommand{\twostepG}[2]{\ensuremath{2\text{-}step_{#1}(#2)}}
\newcommand{\range}[1]{\ensuremath{\mathrm{range}(#1)}}

\newcommand{\Ne}{\mathrm{Ne}}

\definecolor{myyellow}{RGB}{247,192,26}
\definecolor{myblue}{RGB}{37,122,164}
\definecolor{mygreen}{RGB}{78,155,133}
\definecolor{mypurple}{RGB}{86,51,94}

\definecolor{newblue}{RGB}{50,113,173}
\definecolor{newred}{RGB}{222,32,36}
\definecolor{newgreen}{RGB}{70,165,69}
\definecolor{newpurple}{RGB}{140,69,152}

\definecolor{cborange}{RGB}{230,159,0}
\definecolor{cbblue}{RGB}{30,136,229}
\definecolor{cbbluedark}{RGB}{46,37,133}
\definecolor{cbpurple}{RGB}{170,68,153}
\definecolor{cbgreen}{RGB}{0,77,64}
\definecolor{cbgreenlight}{RGB}{93,168,153}
\definecolor{cbbrown}{RGB}{126,41,84}

\pgfdeclarelayer{bg}
\pgfsetlayers{bg,main}

\tikzset{
	rv/.style={draw, ellipse},
	pf/.style={draw, rectangle, draw = black!70, fill = gray!30},
	arc/.style = {->, semithick, >={[round,sep]Stealth}},
	doublearc/.style = {<->, semithick, >={[round,sep]Stealth}},
}

\newcommand\factor[6]{
	\node[pf, #1=#3 of #2, label={#4:{#5}}](#6) {};
}

\newcommand\ufactor[6]{
	\node[pf, pattern=crosshatch, pattern color=gray, #1=#3 of #2, label={#4:{#5}}](#6) {};
}

\newcommand\nodecolorshift[5]{
	\node[circle, fill=#1, above right=0.1em of #2, inner sep=0pt, minimum size=0.5em, xshift=#4, yshift=#5](#3) {};
}

\newcommand\factorcolor[3]{
	\node[circle, fill=#1, above right=0.05em and 0.05em of #2, inner sep=0pt, minimum size=0.5em](#3) {};
}
\newcommand\factorcolorshift[5]{
	\node[circle, fill=#1, above right=0.05em and 0.05em of #2, inner sep=0pt, minimum size=0.5em, xshift=#4, yshift=#5](#3) {};
}

\newcommand\pfs[8]{
	\node[pf, #1=#3 of #2, xshift=-0.1em, yshift=0.1em](#6) {};
	\node[pf, #1=#3 of #2, label={[label distance=0.1em]#4:{#5}}](#7) {};
	\node[pf, #1=#3 of #2, xshift=0.1em, yshift=-0.1em](#8) {};
}
\journal{International Journal of Approximate Reasoning}

\begin{document}
\begin{frontmatter}
	\title{Lifting Factor Graphs with Some Unknown Factors for New Individuals\tnoteref{t1}}
	\tnotetext[t1]{This paper is a revised and extended version of a paper~\citep{Luttermann2023b} that has been published at the Seventeenth European Conference on Symbolic and Quantitative Approaches to Reasoning with Uncertainty (ECSQARU 2023).}
	\author[1]{Malte Luttermann\texorpdfstring{\corref{cor1}}{}}
	\ead{malte.luttermann@dfki.de}
	\cortext[cor1]{Corresponding author}
	\author[2]{Ralf Möller}
	\ead{ralf.moeller@uni-hamburg.de}
	\author[2]{Marcel Gehrke}
	\ead{marcel.gehrke@uni-hamburg.de}
	\address[1]{
		German Research Center for Artificial Intelligence (DFKI),
		Ratzeburger Allee 160,
		23562,
		Lübeck,
		Germany
	}
	\address[2]{
		Institute for Humanities-Centered Artificial Intelligence, University of Hamburg,
		Warburgstraße 28,
		20354,
		Hamburg,
		Germany
	}

	\begin{abstract}
		Lifting exploits symmetries in probabilistic graphical models by using a representative for indistinguishable objects, allowing to carry out query answering more efficiently while maintaining exact answers.
		In this paper, we investigate how lifting enables us to perform probabilistic inference for \aclp{fg} containing unknown factors, i.e., factors whose underlying function of potential mappings is unknown.
		We present the \emph{\ac{lifg} algorithm} to identify indistinguishable subgraphs in a \acl{fg} containing unknown factors, thereby enabling the transfer of known potentials to unknown potentials to ensure a well-defined semantics of the model and allow for (lifted) probabilistic inference.
		We further extend \ac{lifg} to incorporate additional background knowledge about groups of factors belonging to the same individual object.
		By incorporating such background knowledge, \ac{lifg} is able to further reduce the ambiguity of possible transfers of known potentials to unknown potentials.
	\end{abstract}
	\begin{keyword}
		probabilistic graphical models \sep
		factor graphs \sep
		lifted inference
	\end{keyword}
\end{frontmatter}
\acresetall

\section{Introduction}
To perform inference in a probabilistic graphical model, all potential mappings of every factor are required to be known to ensure a well-defined semantics of the model.
However, in practice, scenarios arise in which not all factors are known.
For example, consider a database of a hospital containing patient data and assume a new patient arrives and we want to include them into an existing probabilistic graphical model such as a \ac{fg}.
Clearly, not all attributes included in the database are measured for every new patient, i.e., there are some values missing, resulting in an \ac{fg} with unknown factors and ill-defined semantics when including a new patient in an existing \ac{fg}.
More specifically, it is conceivable that in a first examination of the new patient, a measurement of their blood pressure is conducted whereas measurements for other attributes are not immediately performed.
Therefore, we aim to add new patients to an existing group of indistinguishable patients to treat them equally in the \ac{fg}, thereby allowing for the imputation of missing values under the assumption that there exists such a group for which all values are known.
In particular, we study the problem of constructing a lifted representation having well-defined semantics for an \ac{fg} containing unknown factors---that is, factors whose underlying function mappings from input to output are unknown.
In probabilistic inference, lifting exploits symmetries in a probabilistic graphical model, thereby allowing to carry out query answering more efficiently while maintaining exact answers~\citep{Niepert2014a}.
The main idea behind lifted inference is to use a representative of indistinguishable individuals for computations.
By lifting the probabilistic graphical model, we ensure a well-defined semantics of the model and at the same time allow for tractable probabilistic inference (e.g., inference requiring polynomial time) with respect to domain sizes.

Previous work to construct a lifted representation builds on the \acl{wl} algorithm~\citep{Weisfeiler1968a} which incorporates a colour passing procedure to detect symmetries in a graph, e.g. to test for graph isomorphism.
To construct a lifted representation, denoted as a \ac{pfg}, for a given \ac{fg} in which all factors are known, the \ac{acp} algorithm~\citep{Luttermann2024a,Luttermann2024f,Luttermann2024d} is the state of the art.
The \ac{acp} algorithm builds on the \acl{cp} algorithm (originally named \enquote{CompressFactorGraph})~\citep{Kersting2009a,Ahmadi2013a}, which itself is based on work by \citet{Singla2008a}.
\Ac{acp} detects symmetries in an FG to obtain possible groups of \acp{rv} and factors by deploying a colour passing procedure similar to the \acl{wl} algorithm.
Having obtained a lifted representation, algorithms for lifted inference can be applied.
A widely used algorithm for lifted inference is the \acl{lve} algorithm, first introduced by \citet{Poole2003a} and afterwards refined by many researchers to reach its current form~\citep{DeSalvoBraz2005a,DeSalvoBraz2006a,Milch2008a,Kisynski2009a,Taghipour2013a,Braun2018a}.
Another prominent algorithm for lifted inference is the \acl{ljt} algorithm~\citep{Braun2016a}, which is designed to handle sets of queries instead of single queries.
More recently, causal knowledge has also been incorporated into \acp{pfg} to allow for lifted causal inference~\citep{Luttermann2024b}.

To encounter the problem of constructing a \ac{pfg} as a lifted representation for an \ac{fg} containing unknown factors, we introduce the \ac{lifg} algorithm, which is a generalisation of the \ac{acp} algorithm.
\Ac{lifg} is able to handle arbitrary \acp{fg}, regardless of whether all factors are known or not. By detecting symmetries in an \ac{fg} containing unknown factors, \ac{lifg} generates the possibility to transfer the potentials of known factors to unknown factors to eliminate unknown factors from an \ac{fg}.
We show that, under the assumption that for every unknown factor there is at least one known factor such that they have an indistinguishable surrounding graph structure, \emph{all} unknown potential mappings in an \ac{fg} can be replaced by known potential mappings.
Thereby, \ac{lifg} ensures a well-defined semantics of the model and allows for lifted probabilistic inference.
We further extend \ac{lifg} to incorporate background knowledge about multiple factors belonging to the same individual object---that is, if we know that a set of factors belongs to the same individual object, \ac{lifg} might be able to exploit this knowledge to reduce the ambiguity for possible transfers of known potential mappings.

The remaining part of this paper is structured as follows.
\Cref{sec:prelim} introduces necessary background information and notations. We first recapitulate \acp{fg}, afterwards define \acp{pfg} as first-order probabilistic models, and then describe the \ac{acp} algorithm as a foundation for \ac{lifg}.
Afterwards, in \cref{sec:lifg}, we introduce \ac{lifg} as a generalisation of \ac{acp} allowing us to obtain a lifted representation (a \ac{pfg}) for an \ac{fg} that possibly contains unknown factors.
In \cref{sec:lifg_bk}, we extend \ac{lifg} to incorporate background knowledge.
We then present the results of our empirical evaluation in \cref{sec:eval} before we conclude in \cref{sec:conclusion}.

\section{Preliminaries} \label{sec:prelim}
In this section, we begin by defining \acp{fg} as propositional representations for a joint probability distribution between \acp{rv} and then introduce \acp{pfg}, which combine probabilistic models and first-order logic.
Thereafter, we describe the \ac{acp} algorithm to lift a propositional model, i.e., to transform an \ac{fg} into a \ac{pfg} with equivalent semantics.

\subsection{Factor Graphs and Parameterised Factor Graphs}
An \ac{fg} is an undirected graphical model to compactly encode a full joint probability distribution over a set of \acp{rv} by representing the distribution as a product of factors~\citep{Frey1997a,Kschischang2001a}.

\begin{definition}[Factor Graph, \citealp{Kschischang2001a}] \label{def:lifagu_fg}
	An \emph{\ac{fg}} $G = (\boldsymbol V, \boldsymbol E)$ is an undirected bipartite graph consisting of a node set $\boldsymbol V = \boldsymbol R \cup \boldsymbol F$, where $\boldsymbol R = \{R_1, \ldots, R_n\}$ is a set of \acp{rv} (also referred to as variable nodes) and $\boldsymbol F = \{f_1, \ldots, f_m\}$ is a set of factor nodes, as well as a set of edges $\boldsymbol E \subseteq \boldsymbol R \times \boldsymbol F$.
	Every factor node $f_j \in \boldsymbol F$ defines a function $\phi_j(\mathcal R_j)$, where $\phi_j \colon \times_{R \in \mathcal R_j} \range{R} \mapsto \mathbb{R}^+$ maps a sequence $\mathcal R_j$ of \acp{rv} from $\boldsymbol R$ to a positive real number (called potential).
	The term $\range{R_i}$ denotes the possible values of a \ac{rv} $R_i$.
	There is an edge between a variable node $R_i$ and a factor node $f_j = \phi_j(\mathcal R_j)$ in $\boldsymbol E$ if $R_i$ appears in the argument list of $\phi_j$.
	The semantics of the \ac{fg} $G$ is given by
	\begin{align}
		P_G = \frac{1}{Z} \prod_{j=1}^m \phi_j(\mathcal R_j)
	\end{align}
	with $Z$ being the normalisation constant and $\mathcal R_j$ denoting the \acp{rv} appearing in the argument list of $\phi_j$.
\end{definition}

\begin{figure}[t]
	\centering
	\resizebox{\textwidth}{!}{\begin{tikzpicture}[
	rv/.append style={fill=white}
]
	\node[rv, draw=newred] (E) {$Epid$};

	\factor{above}{E}{0.2cm}{90}{$f_0$}{F0}

	\factor{left}{E}{2cm}{90}{$f_1$}{F1_1}
	\factor{right}{E}{2cm}{90}{$f_1$}{F1_2}

	\node[rv, draw=newblue, below = 1.2cm of F1_1] (SickA) {$Sick.alice$};
	\node[rv, draw=newblue, below = 1.2cm of F1_2] (SickB) {$Sick.bob$};

	\factor{right}{SickA}{0.5cm}{270}{$f_2$}{F2_1}
	\factor{below right}{F2_1}{0.6cm and 0.2cm}{270}{$f_2$}{F2_2}

	\factor{left}{SickB}{0.5cm}{270}{$f_2$}{F2_3}
	\factor{below left}{F2_3}{0.6cm and 0.2cm}{270}{$f_2$}{F2_4}

	\node[rv, draw=newpurple, below left = 0.3cm and 0.4cm of F1_1] (TravelA) {$Travel.alice$};
	\node[rv, draw=newpurple, below right = 0.3cm and 0.4cm of F1_2] (TravelB) {$Travel.bob$};
	\node[rv, draw=newgreen, below left = 0.6cm and -0.8cm of SickA] (TreatAM1) {$Treat.alice.m_1$};
	\node[rv, draw=newgreen, below right = 0.3cm and -0.4cm of TreatAM1] (TreatAM2) {$Treat.alice.m_2$};
	\node[rv, draw=newgreen, below right = 0.6cm and -0.8cm of SickB] (TreatBM1) {$Treat.bob.m_1$};
	\node[rv, draw=newgreen, below left = 0.3cm and -0.4cm of TreatBM1] (TreatBM2) {$Treat.bob.m_2$};

	\factor{left}{TravelA}{0.2cm}{270}{$f_3$}{F3_1}
	\factor{right}{TravelB}{0.2cm}{270}{$f_3$}{F3_2}

	\begin{pgfonlayer}{bg}
		\draw[newred] (E) -- (F0);
		\draw[newred] (E) -- (F1_1);
		\draw[newred] (E) -- (F2_1);
		\draw[newred] (E) -- (F2_2);
		\draw[newred] (E) -- (F1_2);
		\draw[newred] (E) -- (F2_3);
		\draw[newred] (E) -- (F2_4);
		\draw[newblue] (SickA) -- (F1_1);
		\draw[newblue] (SickA) -- (F2_1);
		\draw[newblue] (SickA) -- (F2_2);
		\draw[newpurple] (TravelA) -- (F1_1);
		\draw[newgreen] (TreatAM1.east) -- (F2_1);
		\draw[newgreen] (TreatAM2) -- (F2_2);
		\draw[newblue] (SickB) -- (F1_2);
		\draw[newblue] (SickB) -- (F2_3);
		\draw[newblue] (SickB) -- (F2_4);
		\draw[newpurple] (TravelB) -- (F1_2);
		\draw[newgreen] (TreatBM1.west) -- (F2_3);
		\draw[newgreen] (TreatBM2) -- (F2_4);
		\draw[newpurple] (TravelA) -- (F3_1);
		\draw[newpurple] (TravelB) -- (F3_2);
	\end{pgfonlayer}
\end{tikzpicture}}
	\caption{An \ac{fg} for an epidemic example~\citep{Hoffmann2022a} with two individuals $alice$ and $bob$ as well as two medications $m_1$ and $m_2$ for treatment. For simplicity, all \acp{rv} are Boolean and the input-output pairs of the factors are omitted.}
	\label{fig:epid_fg}
\end{figure}
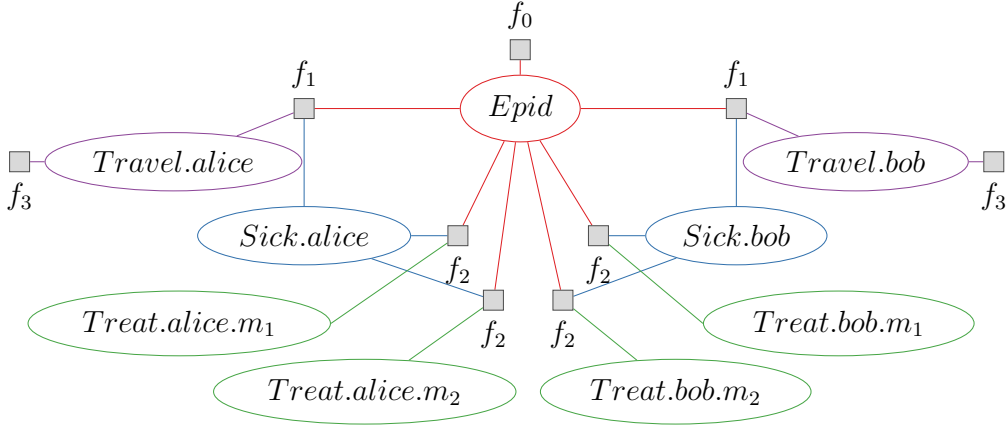

\begin{example}
	\Cref{fig:epid_fg} shows an \ac{fg} representing an epidemic example with two individuals ($alice$ and $bob$) as well as two possible medications ($m_1$ and $m_2$) for treatment.
	For each individual, there are two Boolean \acp{rv} $Sick$ and $Travel$, indicating whether the individual is sick and travels, respectively.
	There is another Boolean \ac{rv} $Treat$ for each combination of individual and medication, specifying whether the individual is treated with the medication.
	The Boolean \ac{rv} $Epid$ states whether an epidemic is present.
	Every factor $f_j$ defines a function, e.g., $f_0 = \phi_0(Epid)$ defines two potential mappings $\phi_0(Epid = true)$ and $\phi_0(Epid = false)$, which are both mapped to a positive real number, respectively.
	We omit the exact specification of the potential mappings from arguments to positive real numbers for brevity.
\end{example}

Note that even though the labelling of the nodes in \cref{fig:epid_fg} may suggest so, there is no explicit representation of individuals in the graph structure of the propositional \ac{fg}.
The labels of the nodes only serve for the reader's understanding and in general, the node labels can be arbitrary strings of characters.
For example, consider the label $Treat.alice.m_1$.
We deliberately avoid using a notation with parameters such as $Treat(alice, m_1)$ to emphasise that the label does \emph{not} explicitly encode that there is an individual $alice$ and a medication $m_1$.
Instead, the label is an arbitrary string of characters (and is only chosen as $Treat.alice.m_1$ for better readability).

Clearly, the size of the \ac{fg} increases with an increasing number of individuals even though it is not necessary to distinguish between individuals because there are symmetries in the model (the factors $f_1$ and $f_3$ occur two times and the factor $f_2$ occurs four times).
In other words, the probability of an epidemic does not depend on knowing which specific individuals are being sick, but only on how many individuals are being sick.
To exploit such symmetries in a model, \acp{pfg} can be used.
In the following, we define \acp{pfg}, first introduced by \citet{Poole2003a}, based on the definitions given by \citet{Gehrke2020a}.
\Acp{pfg} combine first-order logic with probabilistic models, using \acp{lv} as parameters in \acp{rv} to represent sets of indistinguishable \acp{rv}, forming \acp{prv}.

\begin{definition}[Parameterised Random Variable]
	Let $\boldsymbol{R}$ be a set of \ac{rv} names, $\boldsymbol{L}$ a set of \ac{lv} names, $\boldsymbol \Phi$ a set of factor names, and $\boldsymbol{D}$ a set of constants.
	All sets are finite.
	Each \ac{lv} $L$ has a domain $\domain{L} \subseteq \boldsymbol{D}$.
	A \emph{constraint} is a tuple $(\mathcal{X}, \boldsymbol C_{\mathcal{X}})$ of a sequence of \acp{lv} $\mathcal{X} = (X_1, \dots, X_n)$ and a set $\boldsymbol C_{\mathcal{X}} \subseteq \times_{i = 1}^n\domain{X_i}$.
	The symbol $\top$ for $C$ marks that no restrictions apply, i.e., $\boldsymbol C_{\mathcal{X}} = \times_{i = 1}^n\domain{X_i}$.
	A \emph{\ac{prv}} $R(L_1, \dots, L_n)$, $n \geq 0$, is a syntactical construct of a \ac{rv} $R \in \boldsymbol{R}$ possibly combined with \acp{lv} $L_1, \dots, L_n \in \boldsymbol{L}$ to represent a set of \acp{rv}.
	If $n = 0$, the \ac{prv} is parameterless and forms a propositional \ac{rv}.
	A \ac{prv} $A$ (or \ac{lv} $L$) under constraint $C$ is given by $A_{|C}$ ($L_{|C}$, respectively).
	We may omit $|\top$ in $A_{|\top}$ or $L_{|\top}$.
	The term $\range{A}$ denotes the possible values of a \ac{prv} $A$. 
	An \emph{event} $A = a$ denotes the occurrence of \ac{prv} $A$ with range value $a \in \range{A}$ and we call a set of events $\boldsymbol E = \{A_1 = a_1, \dots, A_k = a_k\}$ \emph{evidence}.
\end{definition}

\begin{example} \label{ex:lifagu_prv}
	Consider $\boldsymbol{R} = \{Epid, Travel, Sick, Treat\}$ and $\boldsymbol{L} = \{X,M\}$ with $\domain{X} = \{alice, bob\}$ (people), $\domain{M} = \{m_1, m_2\}$ (medications), combined into Boolean \acp{prv} $Epid$, $Travel(X)$, $Sick(X)$, and $Treat(X,M)$.
\end{example}

In the previous example, there are two indistinguishable individuals $alice$ and $bob$ as well as two indistinguishable medications $m_1$ and $m_2$.
Note that in general, there might be multiple groups of indistinguishable individuals, e.g., in addition to $alice$ and $bob$ there might be another group of individuals $dave$ and $eve$ such that $alice$ and $bob$ as well as $dave$ and $eve$ are indistinguishable, respectively.
To represent multiple groups of indistinguishable objects, constraints are used.
For example, instead of having a single \ac{prv} $Sick(X)$, we might have two \acp{prv} $Sick(X)_{|C_1}$ and $Sick(X)_{|C_2}$ with constraints $C_1 = (X, \{alice, bob\})$ and $C_2 = (X, \{dave, eve\})$, respectively, to represent that $alice$ and $bob$ as well as $dave$ and $eve$ are indistinguishable with respect to being sick.
Analogously, we might have constraints for the \acp{prv} $Travel(X)$ and $Treat(X,M)$, allowing us to represent different combinations of groups of indistinguishable objects.

A \ac{pf} describes a function, mapping argument values to positive real numbers (potentials), of which at least one is non-zero.

\begin{definition}[Parfactor]
	Let $\boldsymbol \Phi$ denote a set of factor names.
	We denote a \emph{\ac{pf}} $g$ by $\phi(\mathcal{A})_{| C}$ with $\mathcal{A} = (A_1, \ldots, A_n)$ being a sequence of \acp{prv}, $\phi \colon \times_{i = 1}^n \range{A_i} \mapsto \mathbb{R}^+$ being a function with name $\phi \in \boldsymbol \Phi$ mapping argument values to a positive real number called \emph{potential}, and $C$ being a constraint on the \acp{lv} of $\mathcal{A}$.
	We may omit $|\top$ in $\phi(\mathcal{A})_{|\top}$.
	The term $lv(Y)$ refers to the \acp{lv} in some element $Y$, a \ac{prv}, a \ac{pf}, or sets thereof.
	The term $gr(Y_{| C})$ denotes the set of all instances (groundings) of $Y$ with respect to constraint $C$.
\end{definition}

\begin{example}
	Take a look at $g_1 = \phi_1(Epid, Travel(X), Sick(X))_{|\top}$ and let $\domain{X} = \{alice, bob\}$.
	If all \acp{prv} are Boolean, $g_1$ specifies $2 \cdot 2 \cdot 2 = 8$ input-output pairs $\phi_1(true, \allowbreak true, \allowbreak true) = \varphi_1$, $\phi_1(true, \allowbreak true, \allowbreak false) = \varphi_2$, and so on with $\varphi_i \in \mathbb{R}^+$.
	Moreover, it holds that $lv(g_1) = \{X\}$ and $gr(g_1) = \{ \phi_1(Epid, \allowbreak Travel(alice), \allowbreak Sick(alice)), \allowbreak \phi_1(Epid, \allowbreak Travel(bob), \allowbreak Sick(bob)) \}$.
	Thus, in this specific example, $g_1$ represents a set of two ground factors.
\end{example}

A set of \acp{pf} $\{g_j\}_{j=1}^m$ then forms a \ac{pfg} $G$.

\begin{definition}[Parametric Factor Graph]
	A \emph{\ac{pfg}} $G = (\boldsymbol V, \boldsymbol E)$ is a bipartite graph with node set $\boldsymbol V = \boldsymbol A \cup \boldsymbol G$ where $\boldsymbol A = \{A_1, \ldots, A_n\}$ is a set of \acp{prv} and $\boldsymbol G = \{g_1, \ldots, g_m\}$ is a set of \acp{pf}.
	A \ac{prv} $A_i$ and a \ac{pf} $g_j$ are connected via an edge in $G$ (i.e., $\{A_i, g_j\} \in \boldsymbol E$) if $A_i$ appears in the argument list of $g_j = \phi_j(\mathcal A_j)$.
	The semantics of $G$ is given by grounding and building a full joint distribution.
	With $Z$ as the normalisation constant and $\mathcal A_k$ denoting the \acp{prv} occurring in the argument list of $\phi_k$, $G$ represents
	\begin{align}
		P_G = \frac{1}{Z} \prod_{g_j \in \boldsymbol G} \prod_{\phi_k \in gr(g_j)} \phi_k(\mathcal A_k).
	\end{align}
\end{definition}

\begin{figure}[t]
	\centering
	\begin{tikzpicture}[
]
	\node[rv, draw=newred] (E) {$Epid$};
	\node[rv, draw=newblue, below = 0.8cm of E] (S) {$Sick(X)$};
	\node[rv, draw=newpurple, left = 0.3cm of S] (Travel) {$Travel(X)$};
	\node[rv, draw=newgreen, right = 0.3cm of S] (Treat) {$Treat(X,M)$};
	\factor{above}{E}{0.2cm}{90}{$g_0$}{G0}
	\pfs{above}{Travel}{0.5cm}{90}{$g_1$}{G1a}{G1}{G1b}
	\pfs{above}{Treat}{0.5cm}{90}{$g_2$}{G2a}{G2}{G2b}
	\pfs{left}{Travel}{0.5cm}{90}{$g_3$}{G3a}{G3}{G3b}

	\begin{pgfonlayer}{bg}
		\draw[newred] (E) -- (G0);
		\draw[newred] (E) -- (G1);
		\draw[newred] (E) -- (G2);
		\draw[newblue] (S) -- (G1);
		\draw[newblue] (S) -- (G2);
		\draw[newpurple] (Travel) -- (G1);
		\draw[newpurple] (Travel) -- (G3);
		\draw[newgreen] (Treat) -- (G2);
	\end{pgfonlayer}
\end{tikzpicture}
	\caption{A \ac{pfg} corresponding to the lifted representation of the \ac{fg} depicted in \cref{fig:epid_fg}. The input-output pairs of the \acp{pf} are again omitted for brevity.}
	\label{fig:epid_pm}
\end{figure}
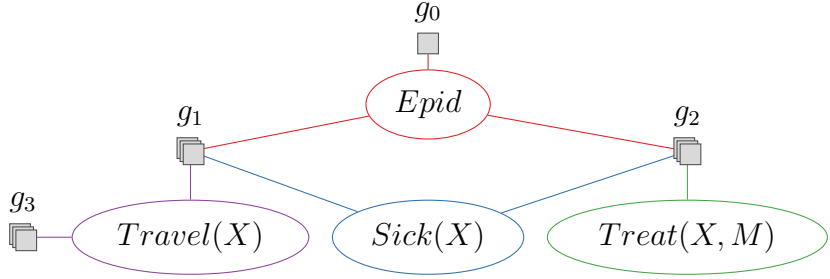

\begin{example}
	\Cref{fig:epid_pm} shows a \ac{pfg} $G$ consisting of the \acp{prv} $Epid$, $Travel(X)$, $Sick(X)$, and $Treat(X,M)$ as well as of the four \acp{pf} $\{g_i\}^3_{i=0}$ where $g_0 = \phi_0(Epid)_{| \top}$, $g_1 = \allowbreak \phi_1(Epid, \allowbreak Travel(X), \allowbreak Sick(X))_{| \top}$, $g_2 = \allowbreak \phi_2(Treat(X,M), \allowbreak Sick(X), \allowbreak Epid)_{| \top}$, and $g_3 = \allowbreak \phi_3(Travel(X), \allowbreak Sick(X))_{| \top}$.
	$G$ is a lifted representation of the \ac{fg} shown in \cref{fig:epid_fg} and the number of \acp{prv} and \acp{pf} in $G$ remains constant even if the number of individuals and medications in the model increases.
\end{example}

We remark that the definition of \acp{pfg} also includes \acp{fg}, as every \ac{fg} is a \ac{pfg} containing only parameterless \acp{rv}.
In \cref{def:lifagu_fg}, we assume that all functions encoded by the factors are known.
As the semantics of an \ac{fg} is given by a product over its factors, the input-output mappings of the factors must be known to ensure a well-defined semantics of the model.
However, in practice, the underlying function specifications of factors might be unknown, leading to the presence of unknown factors in an \ac{fg}.
The upcoming definition formalises the notion of an unknown factor.

\begin{definition}[Unknown Factor]
	Let $G = (\boldsymbol V, \boldsymbol E)$ denote an \ac{fg} with node set $\boldsymbol V = \boldsymbol R \cup \boldsymbol F$, where $\boldsymbol R = \{R_1, \ldots, R_n\}$ is a set of \acp{rv} and $\boldsymbol F = \{f_1, \ldots, f_m\}$ is a set of factors.
	An \emph{unknown factor} $f_j$ is a factor whose function specification $\phi_j(\mathcal R_j)$ is unknown, i.e., the arguments $\mathcal R_j$ of $\phi_j$ are known but the potential values to which $\phi_j$ maps its arguments are unknown.
\end{definition}

Before we deal with unknown factors, we first introduce the \ac{acp} algorithm, which constructs a \ac{pfg} from a given \ac{fg} where all factors are known.
Thereafter, we generalise \ac{acp} to handle the presence unknown factors.

\subsection{The Advanced Colour Passing Algorithm}
The \ac{acp} algorithm~\citep{Luttermann2024a} constructs a lifted representation for an \ac{fg} in which all factors are known.
As \ac{lifg} generalises \ac{acp}, we briefly recap how the \ac{acp} algorithm works.
The idea is to find symmetries in an \ac{fg} based on potentials of factors, ranges and evidence of \acp{rv}, as well as on the graph structure.
Each variable node (\ac{rv}) is assigned a colour depending on its range and observed event, meaning that \acp{rv} with identical ranges and identical observed events are assigned the same colour, and each factor is assigned a colour depending on its potentials, i.e., factors encoding functions with the same potential mappings receive the same colour.
The colours are first passed from every variable node to its neighbouring factor nodes and each factor $f$ collects all colours of neighbouring \acp{rv} in the order of their appearance in the argument list of $f$'s underlying function.
Based on the collected colours and their own colour, factors are grouped together and reassigned a new colour (to reduce communication overhead).
Then, colours are passed from factor nodes to variable nodes.
Based on the collected colours and their own colour, \acp{rv} are grouped together and reassigned a new colour.
The colour passing procedure is iterated until groupings do not change anymore and in the end, all \acp{rv} and factors are grouped together based on their colour signatures (that is, the messages they received from their neighbours plus their own colour).

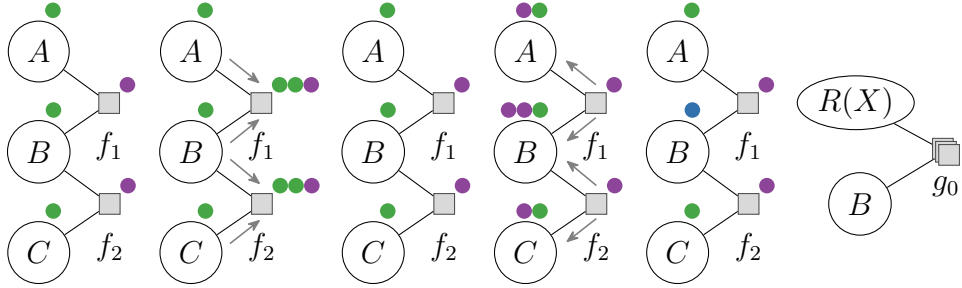
\begin{figure}[t]
	\centering
	\begin{tikzpicture}[label distance=1mm]
	\node[circle, draw] (A) {$A$};
	\node[circle, draw] (B) [below = 0.5cm of A] {$B$};
	\node[circle, draw] (C) [below = 0.5cm of B] {$C$};
	\factor{below right}{A}{0.25cm and 0.5cm}{270}{$f_1$}{f1}
	\factor{below right}{B}{0.25cm and 0.5cm}{270}{$f_2$}{f2}

	\nodecolorshift{newgreen}{A}{Acol}{-2.1mm}{1.5mm}
	\nodecolorshift{newgreen}{B}{Bcol}{-2.1mm}{1.5mm}
	\nodecolorshift{newgreen}{C}{Ccol}{-2.1mm}{1.5mm}

	\factorcolor{newpurple}{f1}{f1col}
	\factorcolor{newpurple}{f2}{f2col}

	\draw (A) -- (f1);
	\draw (B) -- (f1);
	\draw (B) -- (f2);
	\draw (C) -- (f2);

	\node[circle, draw, right = 1.2cm of A] (A1) {$A$};
	\node[circle, draw, below = 0.5cm of A1] (B1) {$B$};
	\node[circle, draw, below = 0.5cm of B1] (C1) {$C$};
	\factor{below right}{A1}{0.25cm and 0.5cm}{270}{$f_1$}{f1_1}
	\factor{below right}{B1}{0.25cm and 0.5cm}{270}{$f_2$}{f2_1}

	\nodecolorshift{newgreen}{A1}{A1col}{-2.1mm}{1.5mm}
	\nodecolorshift{newgreen}{B1}{B1col}{-2.1mm}{1.5mm}
	\nodecolorshift{newgreen}{C1}{C1col}{-2.1mm}{1.5mm}

	\factorcolor{newgreen}{f1_1}{f1_1col1}
	\factorcolorshift{newgreen}{f1_1}{f1_1col2}{2.1mm}{0mm}
	\factorcolorshift{newpurple}{f1_1}{f1_1col3}{4.2mm}{0mm}
	\factorcolor{newgreen}{f2_1}{f2_1col1}
	\factorcolorshift{newgreen}{f2_1}{f2_1col2}{2.1mm}{0mm}
	\factorcolorshift{newpurple}{f2_1}{f2_1col3}{4.2mm}{0mm}

	\coordinate[right=0.1cm of A1, yshift=-0.1cm] (CA1);
	\coordinate[above=0.2cm of f1_1, yshift=-0.1cm] (Cf1_1);
	\coordinate[right=0.1cm of B1, yshift=0.12cm] (CB1);
	\coordinate[right=0.1cm of B1, yshift=-0.1cm] (CB1_1);
	\coordinate[below=0.2cm of f1_1, yshift=0.15cm] (Cf1_1b);
	\coordinate[above=0.2cm of f2_1, yshift=-0.1cm] (Cf2_1);
	\coordinate[right=0.1cm of C1, yshift=0.12cm] (CC1);
	\coordinate[below=0.2cm of f2_1, yshift=0.15cm] (Cf2_1b);

	\begin{pgfonlayer}{bg}
		\draw (A1) -- (f1_1);
		\draw [arc, gray] (CA1) -- (Cf1_1);
		\draw (B1) -- (f1_1);
		\draw [arc, gray] (CB1) -- (Cf1_1b);
		\draw (B1) -- (f2_1);
		\draw [arc, gray] (CB1_1) -- (Cf2_1);
		\draw (C1) -- (f2_1);
		\draw [arc, gray] (CC1) -- (Cf2_1b);
	\end{pgfonlayer}

	\node[circle, draw, right = 1.6cm of A1] (A2) {$A$};
	\node[circle, draw, below = 0.5cm of A2] (B2) {$B$};
	\node[circle, draw, below = 0.5cm of B2] (C2) {$C$};
	\factor{below right}{A2}{0.25cm and 0.5cm}{270}{$f_1$}{f1_2}
	\factor{below right}{B2}{0.25cm and 0.5cm}{270}{$f_2$}{f2_2}

	\nodecolorshift{newgreen}{A2}{A2col}{-2.1mm}{1.5mm}
	\nodecolorshift{newgreen}{B2}{B2col}{-2.1mm}{1.5mm}
	\nodecolorshift{newgreen}{C2}{C2col}{-2.1mm}{1.5mm}

	\factorcolor{newpurple}{f1_2}{f1_2col1}
	\factorcolor{newpurple}{f2_2}{f2_2col1}

	\draw (A2) -- (f1_2);
	\draw (B2) -- (f1_2);
	\draw (B2) -- (f2_2);
	\draw (C2) -- (f2_2);

	\node[circle, draw, right = 1.2cm of A2] (A3) {$A$};
	\node[circle, draw, below = 0.5cm of A3] (B3) {$B$};
	\node[circle, draw, below = 0.5cm of B3] (C3) {$C$};
	\factor{below right}{A3}{0.25cm and 0.5cm}{270}{$f_1$}{f1_3}
	\factor{below right}{B3}{0.25cm and 0.5cm}{270}{$f_2$}{f2_3}

	\nodecolorshift{newpurple}{A3}{A3col1}{-4.2mm}{1.5mm}
	\nodecolorshift{newgreen}{A3}{A3col2}{-2.1mm}{1.5mm}
	\nodecolorshift{newpurple}{B3}{B3col1}{-6.3mm}{1.5mm}
	\nodecolorshift{newpurple}{B3}{B3col2}{-4.2mm}{1.5mm}
	\nodecolorshift{newgreen}{B3}{B3col3}{-2.1mm}{1.5mm}
	\nodecolorshift{newpurple}{C3}{C3col1}{-4.2mm}{1.5mm}
	\nodecolorshift{newgreen}{C3}{C3col2}{-2.1mm}{1.5mm}

	\factorcolor{newpurple}{f1_3}{f1_3col1}
	\factorcolor{newpurple}{f2_3}{f2_3col1}

	\coordinate[right=0.1cm of A3, yshift=-0.1cm] (CA3);
	\coordinate[above=0.2cm of f1_3, yshift=-0.1cm] (Cf1_3);
	\coordinate[right=0.1cm of B3, yshift=0.12cm] (CB3);
	\coordinate[right=0.1cm of B3, yshift=-0.1cm] (CB1_3);
	\coordinate[below=0.2cm of f1_3, yshift=0.15cm] (Cf1_3b);
	\coordinate[above=0.2cm of f2_3, yshift=-0.1cm] (Cf2_3);
	\coordinate[right=0.1cm of C3, yshift=0.12cm] (CC3);
	\coordinate[below=0.2cm of f2_3, yshift=0.15cm] (Cf2_3b);

	\begin{pgfonlayer}{bg}
		\draw (A3) -- (f1_3);
		\draw [arc, gray] (Cf1_3) -- (CA3);
		\draw (B3) -- (f1_3);
		\draw [arc, gray] (Cf1_3b) -- (CB3);
		\draw (B3) -- (f2_3);
		\draw [arc, gray] (Cf2_3) -- (CB1_3);
		\draw (C3) -- (f2_3);
		\draw [arc, gray] (Cf2_3b) -- (CC3);
	\end{pgfonlayer}

	\node[circle, draw, right = 1.2cm of A3] (A4) {$A$};
	\node[circle, draw, below = 0.5cm of A4] (B4) {$B$};
	\node[circle, draw, below = 0.5cm of B4] (C4) {$C$};
	\factor{below right}{A4}{0.25cm and 0.5cm}{270}{$f_1$}{f1_4}
	\factor{below right}{B4}{0.25cm and 0.5cm}{270}{$f_2$}{f2_4}

	\nodecolorshift{newgreen}{A4}{A4col}{-2.1mm}{1.5mm}
	\nodecolorshift{newblue}{B4}{B4col}{-2.1mm}{1.5mm}
	\nodecolorshift{newgreen}{C4}{C4col}{-2.1mm}{1.5mm}

	\factorcolor{newpurple}{f1_4}{f1_4col1}
	\factorcolor{newpurple}{f2_4}{f2_4col1}

	\draw (A4) -- (f1_4);
	\draw (B4) -- (f1_4);
	\draw (B4) -- (f2_4);
	\draw (C4) -- (f2_4);

	\pfs{right}{B4}{3.0cm}{270}{$g_0$}{f12a}{f12}{f12b}

	\node[ellipse, inner sep = 1.2pt, draw, above left = 0.25cm and 0.5cm of f12] (AC) {$R(X)$};
	\node[circle, draw] (B) [below left = 0.25cm and 0.7cm of f12] {$B$};

	\begin{pgfonlayer}{bg}
		\draw (AC) -- (f12);
		\draw (B) -- (f12);
	\end{pgfonlayer}
\end{tikzpicture}
	\caption{The colour passing procedure of the \ac{acp} algorithm on an exemplary input \ac{fg}~\citep{Ahmadi2013a} containing three Boolean \acp{rv} without evidence and two factors encoding functions with identical potential mappings.}
	\label{fig:cp_example}
\end{figure}

\begin{example}
	\Cref{fig:cp_example} depicts the procedure of the \ac{acp} algorithm on a simple \ac{fg}.
	The two factors $f_1$ and $f_2$ encode functions with identical potential mappings in this example.
	As all three \acp{rv} are Boolean and there is no evidence available, $A$, $B$, and $C$ are assigned the same colour (e.g., green).
	Furthermore, the potential mappings of $f_1$ and $f_2$ are identical, so they are assigned the same colour (e.g., purple).
	The colours are then passed from \acp{rv} to factors: $f_1$ receives two times the colour green from $A$ and $B$ and $f_2$ receives two times the colour green from $B$ and $C$.
	Afterwards, $f_1$ and $f_2$ are recoloured according to the colours they received from their neighbours.
	Since both $f_1$ and $f_2$ received the same colours, they are assigned the same colour during recolouring (e.g., purple).
	The colours are then passed from factors to \acp{rv}.
	Here, $A$ receives the colour purple from $f_1$, $B$ receives the colour purple from $f_1$ and the colour purple from $f_2$, and $C$ receives the colour purple from $f_2$.
	Building on these new colour signatures, the \acp{rv} are recoloured such that $A$ and $C$ receive the same colour whereas $B$ is assigned a different colour.
	In this particular example, further iterating the colour passing procedure does not change these groupings.
	Finally, \ac{acp} introduces \acp{lv} to obtain \acp{prv} that represent groups of \acp{rv} with identical colour signatures.
	Further, \ac{acp} replaces groups of factors having identical colours by \acp{pf}.
	Here, $A$ and $C$ are represented by a \ac{prv} $R(X)$ having a single \ac{lv} $X$ with $\domain{X} = \{A,C\}$ (as shown on the right in \cref{fig:cp_example}).
	Note that the name $R$ is chosen arbitrarily and in general, it is also possible to have multiple \acp{lv} within a \ac{prv}.
\end{example}

For more technical details on \ac{acp}, we refer the reader to \citet{Luttermann2024a}.
The authors also provide detailed instructions on how the \acp{lv} are introduced to construct the resulting \ac{pfg} from the colourings.

In a situation with unknown factors being present in an \ac{fg}, the \ac{acp} algorithm cannot be applied to construct a lifted representation for the \ac{fg}.
In the upcoming section, we introduce the \ac{lifg} algorithm which generalises the \ac{acp} algorithm and is able to handle the presence of unknown factors.

\section{The \acs{lifg} Algorithm} \label{sec:lifg}
The semantics of an \ac{fg} (or a \ac{pfg}) relies on a multiplication of all factors in the model and thus, all factors must be known to ensure a well-defined semantics of the model.
As our goal is to perform lifted inference, we have to obtain a \ac{pfg} where all potentials are known.
To transform an \ac{fg} containing unknown factors into a \ac{pfg} without unknown factors, we transfer potentials from known factors to unknown factors.

\begin{figure}[t]
	\centering
	\resizebox{\textwidth}{!}{\begin{tikzpicture}[rv/.append style={fill=white}]
	\node[rv, draw=newred] (E) {$Epid$};

	\factor{above left}{E}{0.2cm and 0.3cm}{[label distance=-1mm]90}{$f_0$}{F0}

	\factor{below left}{E}{-0.1cm and 2cm}{[label distance=-2mm]315}{$f_1$}{F1_1}
	\factor{below right}{E}{-0.1cm and 2cm}{[label distance=-2mm]215}{$f_1$}{F1_2}
	\ufactor{above right}{E}{-0.1cm and 2cm}{[label distance=-2mm]135}{$f_{?}$}{F1_4}

	\node[rv, draw=newblue, below = 0.6cm of F1_1] (SickA) {$Sick.alice$};
	\node[rv, draw=newblue, below = 0.6cm of F1_2] (SickB) {$Sick.bob$};
	\node[rv, draw=newblue, above = 0.6cm of F1_4] (SickE) {$Sick.eve$};

	\factor{right}{SickA}{0.5cm}{[label distance=-1mm]270}{$f_2$}{F2_1}
	\factor{below right}{F2_1}{0.5cm and 0.2cm}{[label distance=-1mm]270}{$f_2$}{F2_2}

	\factor{left}{SickB}{0.5cm}{[label distance=-1mm]270}{$f_2$}{F2_3}
	\factor{below left}{F2_3}{0.5cm and 0.2cm}{[label distance=-1mm]270}{$f_2$}{F2_4}


	\ufactor{left}{SickE}{0.5cm}{[label distance=-1mm]90}{$f_{?}$}{F2_7}
	\ufactor{above left}{F2_7}{0.5cm and 0.2cm}{[label distance=-1mm]90}{$f_{?}$}{F2_8}

	\node[rv, draw=newpurple, below left = 0.0cm and 0.5cm of F1_1] (TravelA) {$Travel.alice$};
	\node[rv, draw=newpurple, below right = 0.0cm and 0.5cm of F1_2] (TravelB) {$Travel.bob$};
	\node[rv, draw=newpurple, above right = 0.0cm and 0.5cm of F1_4] (TravelE) {$Travel.eve$};
	\node[rv, draw=newgreen, below left = 0.2cm and -0.8cm of SickA] (TreatAM1) {$Treat.alice.m_1$};
	\node[rv, draw=newgreen, below right = 0.2cm and -0.5cm of TreatAM1] (TreatAM2) {$Treat.alice.m_2$};
	\node[rv, draw=newgreen, below right = 0.2cm and -0.8cm of SickB] (TreatBM1) {$Treat.bob.m_1$};
	\node[rv, draw=newgreen, below left = 0.2cm and -0.5cm of TreatBM1] (TreatBM2) {$Treat.bob.m_2$};
	\node[rv, draw=newgreen, above right = 0.2cm and -0.8cm of SickE] (TreatEM1) {$Treat.eve.m_1$};
	\node[rv, draw=newgreen, above left = 0.2cm and -0.5cm of TreatEM1] (TreatEM2) {$Treat.eve.m_2$};

	\factor{left}{TravelA}{0.2cm}{270}{$f_3$}{F3_1}
	\factor{right}{TravelB}{0.2cm}{270}{$f_3$}{F3_2}
	\ufactor{right}{TravelE}{0.2cm}{90}{$f_{?}$}{F3_3}

	\begin{pgfonlayer}{bg}
		\draw[newred] (E) -- (F0);
		\draw[newred] (E) -- (F1_1);
		\draw[newred] (E) -- (F2_1);
		\draw[newred] (E) -- (F2_2);
		\draw[newred] (E) -- (F1_2);
		\draw[newred] (E) -- (F2_3);
		\draw[newred] (E) -- (F2_4);
		\draw[newred] (E) -- (F1_4);
		\draw[newred] (E) -- (F2_7);
		\draw[newred] (E) -- (F2_8);
		\draw[newblue] (SickA) -- (F1_1);
		\draw[newblue] (SickA) -- (F2_1);
		\draw[newblue] (SickA) -- (F2_2);
		\draw[newpurple] (TravelA) -- (F1_1);
		\draw[newgreen] (TreatAM1.east) -- (F2_1);
		\draw[newgreen] (TreatAM2) -- (F2_2);
		\draw[newblue] (SickB) -- (F1_2);
		\draw[newblue] (SickB) -- (F2_3);
		\draw[newblue] (SickB) -- (F2_4);
		\draw[newpurple] (TravelB) -- (F1_2);
		\draw[newgreen] (TreatBM1.west) -- (F2_3);
		\draw[newgreen] (TreatBM2) -- (F2_4);
		\draw[newblue] (SickE) -- (F1_4);
		\draw[newblue] (SickE) -- (F2_7);
		\draw[newblue] (SickE) -- (F2_8);
		\draw[newpurple] (TravelE) -- (F1_4);
		\draw[newgreen] (TreatEM1.west) -- (F2_7);
		\draw[newgreen] (TreatEM2) -- (F2_8);
		\draw[newpurple] (TravelA) -- (F3_1);
		\draw[newpurple] (TravelB) -- (F3_2);
		\draw[newpurple] (TravelE) -- (F3_3);
	\end{pgfonlayer}
\end{tikzpicture}}
	\caption{An extension of the epidemic example depicted in \cref{fig:epid_fg}. The factors $f_{?}$ are unknown. The input-output pairs of the remaining factors are again omitted for brevity.}
	\label{fig:epid_fg_extended_01}
\end{figure}

We illustrate the idea of transferring potentials using the exemplary \ac{fg} depicted in \cref{fig:epid_fg_extended_01}.
In this example, another individual $eve$ is added to the model.
Like $alice$ and $bob$, $eve$ can travel, be sick, and be treated and hence, four new \acp{rv} with three new corresponding factors are attached to the model.
However, as we might have limited data, we do not always know the exact potential mappings for the newly introduced factors when a new individual is added to the model and thus, we end up with a model containing unknown factors.
In the example from \cref{fig:epid_fg_extended_01}, we have three unknown factors, denoted as $f_{?}$.
We can transfer the potentials of the known factors $f_1$, $f_2$, and $f_3$ to the newly introduced unknown factors $f_{?}$, as it is reasonable to assume that $eve$ behaves the same as $alice$ and $bob$ as long as no evidence suggesting the contrary is available.

In an \ac{fg} containing unknown factors, the only information available to measure the similarity of factors is the neighbouring graph structure of the factors.
For the upcoming definitions, let $\Ne_G(v)$ denote the set of neighbours of a node $v$ (where $v$ can be a variable node or a factor node) in $G$, i.e., $\Ne_G(f)$ contains all \acp{rv} connected to a factor $f$ in $G$ and $\Ne_G(R)$ contains all factors connected to a \ac{rv} $R$ in $G$.
If the context is clear, we omit the subscript from $\Ne_G(v)$ and write $\Ne(v)$ for simplification.
We start by defining the 2-step neighbourhood of a factor $f$ as the set containing $f$, all \acp{rv} that are connected to $f$, as well as all factors connected to a \ac{rv} that is connected to $f$.
The concept of taking into account all nodes with a maximal distance of two is based on the idea of considering a single iteration of the colour passing procedure in the \ac{acp} algorithm.

\begin{definition}[2-Step Neighbourhood]
	The \emph{2-step neighbourhood} of a factor $f$ in an \ac{fg} $G$ is defined as
	\begin{align*}
		\twostepG{G}{f} = \{ R \mid R \in \Ne_G(f) \} \cup \{ f' \mid \exists R \colon R \in \Ne_G(f) \land f' \in \Ne_G(R) \}.
	\end{align*}
\end{definition}

If the context is clear, we write \twostep{f} instead of \twostepG{G}{f}.

\begin{example}
	The 2-step neighbourhood of $f_1$ in the \ac{fg} depicted in \cref{fig:cp_example} is given by $\twostep{f_1} = \{A,B\} \cup \{f_1, f_2\}$.
	By $G[V']$ we denote the subgraph of a graph $G$ induced by a subset of nodes $V'$, that is, $G[V']$ contains only the nodes in $V'$ as well as all edges from $G$ that connect two nodes in $V'$.
	In this example, $G[\twostep{f_1}]$ then consists of the nodes $A$, $B$, $f_1$, and $f_2$, and contains the edges $A - f_1$, $B - f_1$, and $B - f_2$.
\end{example}

As it is currently unknown whether a general graph isomorphism test is solvable in polynomial time, we make use of the weaker notion of indistinguishable 2-step neighbourhoods instead of relying on isomorphic 2-step neighbourhoods to ensure that \ac{lifg} is implementable in polynomial time.

\begin{definition}[Indistinguishable 2-Step Neighbourhoods] \label{def:lifagu_indistinguishable_two_step}
	Let $G$ denote an \ac{fg} and let $f_i$ as well as $f_j$ denote two factors in $G$.
	Then, $G[\twostepG{G}{f_i}]$ and $G[\twostepG{G}{f_j}]$ are \emph{indistinguishable} if
	\begin{enumerate}
		\item \label{itm:lifagu_indistinguishable_two_step_1} $\abs{\Ne_G(f_i)} = \abs{\Ne_G(f_j)}$ and
		\item \label{itm:lifagu_indistinguishable_two_step_2} there exists a bijection $\tau \colon \Ne_G(f_i) \mapsto \Ne_G(f_j)$ that maps every \ac{rv} $R_k \in \Ne_G(f_i)$ to a \ac{rv} $R_{\ell} \in \Ne_G(f_j)$ such that the observed event (evidence) for $R_k$ and $R_{\ell}$ is identical, $\range{R_k} = \range{R_{\ell}}$, and $\abs{\Ne_G(R_k)} = \abs{\Ne_G(R_{\ell})}$.
	\end{enumerate}
\end{definition}

\begin{example}
	Take a look again at the \ac{fg} shown in \cref{fig:cp_example} and assume that there is no evidence.
	We can check whether $f_1$ and $f_2$ have indistinguishable 2-step neighbourhoods: Both $f_1$ and $f_2$ are connected to two \acp{rv} as $\Ne(f_1) = \{A,B\}$ and $\Ne(f_2) = \{B,C\}$, thereby satisfying \cref{itm:lifagu_indistinguishable_two_step_1}.
	Further, $A$ can be mapped to $C$ with $\range{A} = \range{C}$ (Boolean) and $\abs{\Ne(A)} = \abs{\Ne(C)} = 1$ and $B$ can be mapped to itself.
	Thus, \cref{itm:lifagu_indistinguishable_two_step_2} is satisfied and it holds that $G[\twostep{f_1}]$ and $G[\twostep{f_2}]$ are indistinguishable.
\end{example}

Recall that our goal is to transfer potentials from known factors to unknown factors.
To do so, we need a measure of similarity between factors, even if the underlying functions encoded by the factors are unknown.
We thus introduce the notion of possibly identical factors, that is, factors that are indistinguishable based on their 2-step neighbourhoods.
In particular, two factors are considered possibly identical if the subgraphs induced by their 2-step neighbourhoods are indistinguishable and the underlying functions (if known) do not differ from each other, as formalised in the next definition.

\begin{definition}[Possibly Identical Factors] \label{def:lifagu_possibly_identical}
	Given two factors $f_i$ and $f_j$ in an \ac{fg} $G$, we call $f_i$ and $f_j$ \emph{possibly identical}, denoted as $f_i \approx f_j$, if
	\begin{enumerate}
		\item \label{itm:lifagu_possibly_identical_itm_1} $G[\twostepG{G}{f_i}]$ and $G[\twostepG{G}{f_j}]$ are indistinguishable, and
		\item \label{itm:lifagu_possibly_identical_itm_2} at least one of $f_i$, $f_j$ is unknown, or the underlying functions of $f_i$ and $f_j$ encode identical potential mappings.
	\end{enumerate}
\end{definition}

\Cref{itm:lifagu_possibly_identical_itm_2} serves to ensure consistency when comparing factors with known underlying functions as two factors encoding different potential mappings can obviously not be identical.

\begin{example}
	Applying the definition of possibly identical factors to $f_1$ and $f_2$ from \cref{fig:cp_example}, we can verify that $f_1$ and $f_2$ are indeed possibly identical because they have indistinguishable 2-step neighbourhoods and their underlying functions encode identical potential mappings.
\end{example}

We are now ready to introduce the \ac{lifg} algorithm, which makes use of the notion of possibly identical factors to find known factors that are similar to unknown factors.
\Cref{alg:lifg} outlines the entire \ac{lifg} algorithm and we provide a detailed explanation of each step in the following.

\begin{algorithm}[t]
	\SetKwInOut{Input}{Input}
	\SetKwInOut{Output}{Output}
	\caption{\ac{lifg}}
	\label{alg:lifg}
	\Input{An \ac{fg} $G$ with \acp{rv} $\boldsymbol{R} = \{R_1, \dots, R_n\}$, known factors $\boldsymbol{F} = \{f_1, \dots, f_m\}$, unknown factors $\boldsymbol{F'} = \{f'_1, \dots, f'_z\}$, and evidence $\boldsymbol{E} = \{R_1 = r_1, \dots, R_k = r_k\}$, as well as a real-valued threshold $\theta \in [0, 1]$.}
	\Output{A lifted representation $G'$ of $G$.}
	\BlankLine
	Assign each $f_i \in \boldsymbol{F}$ a colour based on its potentials\; \label{line:assign_known_colours}
	Assign each $f'_i \in \boldsymbol{F'}$ a unique colour\; \label{line:assign_unknown_colours}
	\ForEach{unknown factor $f_i \in \boldsymbol{F'}$}{
		$C_{f_i} \gets \{\}$\;
		\ForEach{factor $f_j \in \boldsymbol{F} \cup \boldsymbol{F'}$ with $f_i \neq f_j$}{
			\If{$f_i \approx f_j$}{
				\eIf{$f_j$ is unknown}{
					Assign $f_j$ the same colour as $f_i$\; \label{line:reassign_unknown_colours}
				}{
					$C_{f_i} \gets C_{f_i} \cup \{f_j\}$\; \label{line:add_to_candidates}
				}
			}
		}
	}
	\ForEach{set of candidates $C_{f_i}$}{
		$C_{f_i}^{\ell} \gets$ Maximal subset of $C_{f_i}$ s.t.\ $f_j \approx f_k$ holds for all $f_j, f_k \in C_{f_i}^{\ell}$\; \label{line:max_subset_candidates}
		\If{$\abs{C_{f_i}^{\ell}} \mathbin{/} \abs{C_{f_i}} \geq \theta$}{ \label{line:if_threshold}
			Assign all $f_j \in C_{f_i}^{\ell}$ the same colour as $f_i$\; \label{line:assign_unkown_from_known}
			Assign $f_i$ the same potentials as the factors $f_j \in C_{f_i}^{\ell}$\; \label{line:transfer_unkown_from_known}
		}
	}
	$G \gets$ Result from calling \ac{acp} on the modified graph $G$ and $\boldsymbol{E}$\; \label{line:call_cp}
\end{algorithm}

\Ac{lifg} assigns colours to unknown factors based on indistinguishable 2-step neighbourhoods, proceeding as follows for an input $G$.
As an initialisation step, \ac{lifg} assigns each known factor a colour based on its potentials and each unknown factor a unique colour.
Then, \ac{lifg} searches for possibly identical factors in two phases.
In the first phase, all unknown factors that are possibly identical are assigned the same colour, as there is no way to distinguish them.
Furthermore, \ac{lifg} collects for every unknown factor $f_i$ a set $C_{f_i}$ of known factors possibly identical to $f_i$.
The second phase then continues to group the unknown factors with known factors, including the transfer of the potentials from the known factors to the unknown factors.
For every unknown factor $f_i$, \ac{lifg} computes a maximal subset $C_{f_i}^{\ell} \subseteq C_{f_i}$ for which all elements are pairwise possibly identical.
Afterwards, $f_i$ and all $f_j \in C_{f_i}^{\ell}$ are assigned the same colour if a user-defined threshold is reached.
At the same time, the potentials of the factors in $C_{f_i}^{\ell}$ are transferred to $f_i$.
Finally, \ac{acp} is called on $G$, which now includes the previously set colours for the unknown factors in $G$, to group both known and unknown factors.

Before we take a closer look at the threshold $\theta$, we illustrate the steps undertaken by the \ac{lifg} algorithm on the exemplary \ac{fg} from \cref{fig:epid_fg_extended_01}.

\begin{figure}[t]
	\centering
	\begin{subfigure}{0.49\linewidth}
		\centering
		\resizebox{\textwidth}{!}{\begin{tikzpicture}[rv/.append style={fill=white}]
	\node[rv, draw=newred] (E) {$Epid$};

	\factor{above left}{E}{0.2cm and 0.3cm}{[label distance=-1mm]90}{$f_0$}{F0}

	\factor{below left}{E}{-0.1cm and 2cm}{[label distance=-2mm]315}{$f_1$}{F1_1}
	\factor{below right}{E}{-0.1cm and 2cm}{[label distance=-2mm]215}{$f_1$}{F1_2}
	\ufactor{above right}{E}{-0.1cm and 2cm}{[label distance=-2mm]135}{$f_{?}$}{F1_4}

	\node[rv, draw=newblue, below = 0.6cm of F1_1] (SickA) {$Sick.alice$};
	\node[rv, draw=newblue, below = 0.6cm of F1_2] (SickB) {$Sick.bob$};
	\node[rv, draw=newblue, above = 0.6cm of F1_4] (SickE) {$Sick.eve$};

	\factor{right}{SickA}{0.5cm}{[label distance=-1mm]270}{$f_2$}{F2_1}
	\factor{below right}{F2_1}{0.5cm and 0.2cm}{[label distance=-1mm]270}{$f_2$}{F2_2}

	\factor{left}{SickB}{0.5cm}{[label distance=-1mm]270}{$f_2$}{F2_3}
	\factor{below left}{F2_3}{0.5cm and 0.2cm}{[label distance=-1mm]270}{$f_2$}{F2_4}


	\ufactor{left}{SickE}{0.5cm}{[label distance=-1mm]90}{$f_{?}$}{F2_7}
	\ufactor{above left}{F2_7}{0.5cm and 0.2cm}{[label distance=-1mm]90}{$f_{?}$}{F2_8}

	\node[rv, draw=newpurple, below left = 0.0cm and 0.5cm of F1_1] (TravelA) {$Travel.alice$};
	\node[rv, draw=newpurple, below right = 0.0cm and 0.5cm of F1_2] (TravelB) {$Travel.bob$};
	\node[rv, draw=newpurple, above right = 0.0cm and 0.5cm of F1_4] (TravelE) {$Travel.eve$};
	\node[rv, draw=newgreen, below left = 0.2cm and -0.8cm of SickA] (TreatAM1) {$Treat.alice.m_1$};
	\node[rv, draw=newgreen, below right = 0.2cm and -0.5cm of TreatAM1] (TreatAM2) {$Treat.alice.m_2$};
	\node[rv, draw=newgreen, below right = 0.2cm and -0.8cm of SickB] (TreatBM1) {$Treat.bob.m_1$};
	\node[rv, draw=newgreen, below left = 0.2cm and -0.5cm of TreatBM1] (TreatBM2) {$Treat.bob.m_2$};
	\node[rv, draw=newgreen, above right = 0.2cm and -0.8cm of SickE] (TreatEM1) {$Treat.eve.m_1$};
	\node[rv, draw=newgreen, above left = 0.2cm and -0.5cm of TreatEM1] (TreatEM2) {$Treat.eve.m_2$};

	\factor{left}{TravelA}{0.2cm}{270}{$f_3$}{F3_1}
	\factor{right}{TravelB}{0.2cm}{270}{$f_3$}{F3_2}
	\ufactor{right}{TravelE}{0.2cm}{90}{$f_{?}$}{F3_3}

	\factorcolorshift{newred!80}{F0}{F0_col}{0.2mm}{0mm}
	\factorcolorshift{newblue!80}{F1_1}{F1_1_col}{0.2mm}{0mm}
	\factorcolorshift{newblue!80}{F1_2}{F1_2_col}{0.2mm}{0mm}
	\factorcolorshift{newgreen!80}{F2_1}{F2_1_col}{0.2mm}{0mm}
	\factorcolorshift{newgreen!80}{F2_2}{F2_2_col}{0.2mm}{0mm}
	\factorcolorshift{newgreen!80}{F2_3}{F2_3_col}{0.2mm}{0mm}
	\factorcolorshift{newgreen!80}{F2_4}{F2_4_col}{0.2mm}{0mm}
	\factorcolorshift{newpurple!80}{F3_1}{F3_1_col}{0.2mm}{0mm}
	\factorcolorshift{newpurple!80}{F3_2}{F3_2_col}{0.2mm}{0mm}

	\begin{pgfonlayer}{bg}
		\draw[newred] (E) -- (F0);
		\draw[newred] (E) -- (F1_1);
		\draw[newred] (E) -- (F2_1);
		\draw[newred] (E) -- (F2_2);
		\draw[newred] (E) -- (F1_2);
		\draw[newred] (E) -- (F2_3);
		\draw[newred] (E) -- (F2_4);
		\draw[newred] (E) -- (F1_4);
		\draw[newred] (E) -- (F2_7);
		\draw[newred] (E) -- (F2_8);
		\draw[newblue] (SickA) -- (F1_1);
		\draw[newblue] (SickA) -- (F2_1);
		\draw[newblue] (SickA) -- (F2_2);
		\draw[newpurple] (TravelA) -- (F1_1);
		\draw[newgreen] (TreatAM1.east) -- (F2_1);
		\draw[newgreen] (TreatAM2) -- (F2_2);
		\draw[newblue] (SickB) -- (F1_2);
		\draw[newblue] (SickB) -- (F2_3);
		\draw[newblue] (SickB) -- (F2_4);
		\draw[newpurple] (TravelB) -- (F1_2);
		\draw[newgreen] (TreatBM1.west) -- (F2_3);
		\draw[newgreen] (TreatBM2) -- (F2_4);
		\draw[newblue] (SickE) -- (F1_4);
		\draw[newblue] (SickE) -- (F2_7);
		\draw[newblue] (SickE) -- (F2_8);
		\draw[newpurple] (TravelE) -- (F1_4);
		\draw[newgreen] (TreatEM1.west) -- (F2_7);
		\draw[newgreen] (TreatEM2) -- (F2_8);
		\draw[newpurple] (TravelA) -- (F3_1);
		\draw[newpurple] (TravelB) -- (F3_2);
		\draw[newpurple] (TravelE) -- (F3_3);
	\end{pgfonlayer}
\end{tikzpicture}}
		\caption{}
		\label{fig:epid_fg_lifagu_01}
	\end{subfigure}
	\begin{subfigure}{0.49\linewidth}
		\centering
		\resizebox{\textwidth}{!}{\begin{tikzpicture}[rv/.append style={fill=white}]
	\node[rv, draw=newred] (E) {$Epid$};

	\factor{above left}{E}{0.2cm and 0.3cm}{[label distance=-1mm]90}{$f_0$}{F0}

	\factor{below left}{E}{-0.1cm and 2cm}{[label distance=-2mm]315}{$f_1$}{F1_1}
	\factor{below right}{E}{-0.1cm and 2cm}{[label distance=-2mm]215}{$f_1$}{F1_2}
	\ufactor{above right}{E}{-0.1cm and 2cm}{[label distance=-2mm]135}{$f_{?}$}{F1_4}

	\node[rv, draw=newblue, below = 0.6cm of F1_1] (SickA) {$Sick.alice$};
	\node[rv, draw=newblue, below = 0.6cm of F1_2] (SickB) {$Sick.bob$};
	\node[rv, draw=newblue, above = 0.6cm of F1_4] (SickE) {$Sick.eve$};

	\factor{right}{SickA}{0.5cm}{[label distance=-1mm]270}{$f_2$}{F2_1}
	\factor{below right}{F2_1}{0.5cm and 0.2cm}{[label distance=-1mm]270}{$f_2$}{F2_2}

	\factor{left}{SickB}{0.5cm}{[label distance=-1mm]270}{$f_2$}{F2_3}
	\factor{below left}{F2_3}{0.5cm and 0.2cm}{[label distance=-1mm]270}{$f_2$}{F2_4}


	\ufactor{left}{SickE}{0.5cm}{[label distance=-1mm]90}{$f_{?}$}{F2_7}
	\ufactor{above left}{F2_7}{0.5cm and 0.2cm}{[label distance=-1mm]90}{$f_{?}$}{F2_8}

	\node[rv, draw=newpurple, below left = 0.0cm and 0.5cm of F1_1] (TravelA) {$Travel.alice$};
	\node[rv, draw=newpurple, below right = 0.0cm and 0.5cm of F1_2] (TravelB) {$Travel.bob$};
	\node[rv, draw=newpurple, above right = 0.0cm and 0.5cm of F1_4] (TravelE) {$Travel.eve$};
	\node[rv, draw=newgreen, below left = 0.2cm and -0.8cm of SickA] (TreatAM1) {$Treat.alice.m_1$};
	\node[rv, draw=newgreen, below right = 0.2cm and -0.5cm of TreatAM1] (TreatAM2) {$Treat.alice.m_2$};
	\node[rv, draw=newgreen, below right = 0.2cm and -0.8cm of SickB] (TreatBM1) {$Treat.bob.m_1$};
	\node[rv, draw=newgreen, below left = 0.2cm and -0.5cm of TreatBM1] (TreatBM2) {$Treat.bob.m_2$};
	\node[rv, draw=newgreen, above right = 0.2cm and -0.8cm of SickE] (TreatEM1) {$Treat.eve.m_1$};
	\node[rv, draw=newgreen, above left = 0.2cm and -0.5cm of TreatEM1] (TreatEM2) {$Treat.eve.m_2$};

	\factor{left}{TravelA}{0.2cm}{270}{$f_3$}{F3_1}
	\factor{right}{TravelB}{0.2cm}{270}{$f_3$}{F3_2}
	\ufactor{right}{TravelE}{0.2cm}{90}{$f_{?}$}{F3_3}

	\factorcolorshift{newred!80}{F0}{F0_col}{0.2mm}{0mm}
	\factorcolorshift{newblue!80}{F1_1}{F1_1_col}{0.2mm}{0mm}
	\factorcolorshift{newblue!80}{F1_2}{F1_2_col}{0.2mm}{0mm}
	\factorcolorshift{newgreen!80}{F2_1}{F2_1_col}{0.2mm}{0mm}
	\factorcolorshift{newgreen!80}{F2_2}{F2_2_col}{0.2mm}{0mm}
	\factorcolorshift{newgreen!80}{F2_3}{F2_3_col}{0.2mm}{0mm}
	\factorcolorshift{newgreen!80}{F2_4}{F2_4_col}{0.2mm}{0mm}
	\factorcolorshift{newpurple!80}{F3_1}{F3_1_col}{0.2mm}{0mm}
	\factorcolorshift{newpurple!80}{F3_2}{F3_2_col}{0.2mm}{0mm}

	\factorcolorshift{cborange!80}{F1_4}{F1_4_col}{0.2mm}{0mm}
	\factorcolorshift{cbbrown!80}{F2_7}{F2_7_col}{0.2mm}{0mm}
	\factorcolorshift{black!80}{F2_8}{F2_8_col}{0.2mm}{0mm}
	\factorcolorshift{gray!80}{F3_3}{F3_3_col}{0.2mm}{0mm}

	\begin{pgfonlayer}{bg}
		\draw[newred] (E) -- (F0);
		\draw[newred] (E) -- (F1_1);
		\draw[newred] (E) -- (F2_1);
		\draw[newred] (E) -- (F2_2);
		\draw[newred] (E) -- (F1_2);
		\draw[newred] (E) -- (F2_3);
		\draw[newred] (E) -- (F2_4);
		\draw[newred] (E) -- (F1_4);
		\draw[newred] (E) -- (F2_7);
		\draw[newred] (E) -- (F2_8);
		\draw[newblue] (SickA) -- (F1_1);
		\draw[newblue] (SickA) -- (F2_1);
		\draw[newblue] (SickA) -- (F2_2);
		\draw[newpurple] (TravelA) -- (F1_1);
		\draw[newgreen] (TreatAM1.east) -- (F2_1);
		\draw[newgreen] (TreatAM2) -- (F2_2);
		\draw[newblue] (SickB) -- (F1_2);
		\draw[newblue] (SickB) -- (F2_3);
		\draw[newblue] (SickB) -- (F2_4);
		\draw[newpurple] (TravelB) -- (F1_2);
		\draw[newgreen] (TreatBM1.west) -- (F2_3);
		\draw[newgreen] (TreatBM2) -- (F2_4);
		\draw[newblue] (SickE) -- (F1_4);
		\draw[newblue] (SickE) -- (F2_7);
		\draw[newblue] (SickE) -- (F2_8);
		\draw[newpurple] (TravelE) -- (F1_4);
		\draw[newgreen] (TreatEM1.west) -- (F2_7);
		\draw[newgreen] (TreatEM2) -- (F2_8);
		\draw[newpurple] (TravelA) -- (F3_1);
		\draw[newpurple] (TravelB) -- (F3_2);
		\draw[newpurple] (TravelE) -- (F3_3);
	\end{pgfonlayer}
\end{tikzpicture}}
		\caption{}
		\label{fig:epid_fg_lifagu_02}
	\end{subfigure}

	\begin{subfigure}{0.49\linewidth}
		\centering
		\resizebox{\textwidth}{!}{\begin{tikzpicture}[rv/.append style={fill=white}]
	\node[rv, draw=newred] (E) {$Epid$};

	\factor{above left}{E}{0.2cm and 0.3cm}{[label distance=-1mm]90}{$f_0$}{F0}

	\factor{below left}{E}{-0.1cm and 2cm}{[label distance=-2mm]315}{$f_1$}{F1_1}
	\factor{below right}{E}{-0.1cm and 2cm}{[label distance=-2mm]215}{$f_1$}{F1_2}
	\ufactor{above right}{E}{-0.1cm and 2cm}{[label distance=-2mm]135}{$f_{?}$}{F1_4}

	\node[rv, draw=newblue, below = 0.6cm of F1_1] (SickA) {$Sick.alice$};
	\node[rv, draw=newblue, below = 0.6cm of F1_2] (SickB) {$Sick.bob$};
	\node[rv, draw=newblue, above = 0.6cm of F1_4] (SickE) {$Sick.eve$};

	\factor{right}{SickA}{0.5cm}{[label distance=-1mm]270}{$f_2$}{F2_1}
	\factor{below right}{F2_1}{0.5cm and 0.2cm}{[label distance=-1mm]270}{$f_2$}{F2_2}

	\factor{left}{SickB}{0.5cm}{[label distance=-1mm]270}{$f_2$}{F2_3}
	\factor{below left}{F2_3}{0.5cm and 0.2cm}{[label distance=-1mm]270}{$f_2$}{F2_4}


	\ufactor{left}{SickE}{0.5cm}{[label distance=-1mm]90}{$f_{?}$}{F2_7}
	\ufactor{above left}{F2_7}{0.5cm and 0.2cm}{[label distance=-1mm]90}{$f_{?}$}{F2_8}

	\node[rv, draw=newpurple, below left = 0.0cm and 0.5cm of F1_1] (TravelA) {$Travel.alice$};
	\node[rv, draw=newpurple, below right = 0.0cm and 0.5cm of F1_2] (TravelB) {$Travel.bob$};
	\node[rv, draw=newpurple, above right = 0.0cm and 0.5cm of F1_4] (TravelE) {$Travel.eve$};
	\node[rv, draw=newgreen, below left = 0.2cm and -0.8cm of SickA] (TreatAM1) {$Treat.alice.m_1$};
	\node[rv, draw=newgreen, below right = 0.2cm and -0.5cm of TreatAM1] (TreatAM2) {$Treat.alice.m_2$};
	\node[rv, draw=newgreen, below right = 0.2cm and -0.8cm of SickB] (TreatBM1) {$Treat.bob.m_1$};
	\node[rv, draw=newgreen, below left = 0.2cm and -0.5cm of TreatBM1] (TreatBM2) {$Treat.bob.m_2$};
	\node[rv, draw=newgreen, above right = 0.2cm and -0.8cm of SickE] (TreatEM1) {$Treat.eve.m_1$};
	\node[rv, draw=newgreen, above left = 0.2cm and -0.5cm of TreatEM1] (TreatEM2) {$Treat.eve.m_2$};

	\factor{left}{TravelA}{0.2cm}{270}{$f_3$}{F3_1}
	\factor{right}{TravelB}{0.2cm}{270}{$f_3$}{F3_2}
	\ufactor{right}{TravelE}{0.2cm}{90}{$f_{?}$}{F3_3}

	\factorcolorshift{newred!80}{F0}{F0_col}{0.2mm}{0mm}
	\factorcolorshift{newblue!80}{F1_1}{F1_1_col}{0.2mm}{0mm}
	\factorcolorshift{newblue!80}{F1_2}{F1_2_col}{0.2mm}{0mm}
	\factorcolorshift{newgreen!80}{F2_1}{F2_1_col}{0.2mm}{0mm}
	\factorcolorshift{newgreen!80}{F2_2}{F2_2_col}{0.2mm}{0mm}
	\factorcolorshift{newgreen!80}{F2_3}{F2_3_col}{0.2mm}{0mm}
	\factorcolorshift{newgreen!80}{F2_4}{F2_4_col}{0.2mm}{0mm}
	\factorcolorshift{newpurple!80}{F3_1}{F3_1_col}{0.2mm}{0mm}
	\factorcolorshift{newpurple!80}{F3_2}{F3_2_col}{0.2mm}{0mm}

	\factorcolorshift{cborange!80}{F1_4}{F1_4_col}{0.2mm}{0mm}
	\factorcolorshift{cbbrown!80}{F2_7}{F2_7_col}{0.2mm}{0mm}
	\factorcolorshift{cbbrown!80}{F2_8}{F2_8_col}{0.2mm}{0mm}
	\factorcolorshift{gray!80}{F3_3}{F3_3_col}{0.2mm}{0mm}

	\begin{pgfonlayer}{bg}
		\draw[newred] (E) -- (F0);
		\draw[newred] (E) -- (F1_1);
		\draw[newred] (E) -- (F2_1);
		\draw[newred] (E) -- (F2_2);
		\draw[newred] (E) -- (F1_2);
		\draw[newred] (E) -- (F2_3);
		\draw[newred] (E) -- (F2_4);
		\draw[newred] (E) -- (F1_4);
		\draw[newred] (E) -- (F2_7);
		\draw[newred] (E) -- (F2_8);
		\draw[newblue] (SickA) -- (F1_1);
		\draw[newblue] (SickA) -- (F2_1);
		\draw[newblue] (SickA) -- (F2_2);
		\draw[newpurple] (TravelA) -- (F1_1);
		\draw[newgreen] (TreatAM1.east) -- (F2_1);
		\draw[newgreen] (TreatAM2) -- (F2_2);
		\draw[newblue] (SickB) -- (F1_2);
		\draw[newblue] (SickB) -- (F2_3);
		\draw[newblue] (SickB) -- (F2_4);
		\draw[newpurple] (TravelB) -- (F1_2);
		\draw[newgreen] (TreatBM1.west) -- (F2_3);
		\draw[newgreen] (TreatBM2) -- (F2_4);
		\draw[newblue] (SickE) -- (F1_4);
		\draw[newblue] (SickE) -- (F2_7);
		\draw[newblue] (SickE) -- (F2_8);
		\draw[newpurple] (TravelE) -- (F1_4);
		\draw[newgreen] (TreatEM1.west) -- (F2_7);
		\draw[newgreen] (TreatEM2) -- (F2_8);
		\draw[newpurple] (TravelA) -- (F3_1);
		\draw[newpurple] (TravelB) -- (F3_2);
		\draw[newpurple] (TravelE) -- (F3_3);
	\end{pgfonlayer}
\end{tikzpicture}}
		\caption{}
		\label{fig:epid_fg_lifagu_03}
	\end{subfigure}
	\begin{subfigure}{0.49\linewidth}
		\centering
		\resizebox{\textwidth}{!}{\begin{tikzpicture}[rv/.append style={fill=white}]
	\node[rv, draw=newred] (E) {$Epid$};

	\factor{above left}{E}{0.2cm and 0.3cm}{[label distance=-1mm]90}{$f_0$}{F0}

	\factor{below left}{E}{-0.1cm and 2cm}{[label distance=-2mm]315}{$f_1$}{F1_1}
	\factor{below right}{E}{-0.1cm and 2cm}{[label distance=-2mm]215}{$f_1$}{F1_2}
	\factor{above right}{E}{-0.1cm and 2cm}{[label distance=-2mm]135}{$f_1$}{F1_4}

	\node[rv, draw=newblue, below = 0.6cm of F1_1] (SickA) {$Sick.alice$};
	\node[rv, draw=newblue, below = 0.6cm of F1_2] (SickB) {$Sick.bob$};
	\node[rv, draw=newblue, above = 0.6cm of F1_4] (SickE) {$Sick.eve$};

	\factor{right}{SickA}{0.5cm}{[label distance=-1mm]270}{$f_2$}{F2_1}
	\factor{below right}{F2_1}{0.5cm and 0.2cm}{[label distance=-1mm]270}{$f_2$}{F2_2}

	\factor{left}{SickB}{0.5cm}{[label distance=-1mm]270}{$f_2$}{F2_3}
	\factor{below left}{F2_3}{0.5cm and 0.2cm}{[label distance=-1mm]270}{$f_2$}{F2_4}


	\factor{left}{SickE}{0.5cm}{[label distance=-1mm]90}{$f_2$}{F2_7}
	\factor{above left}{F2_7}{0.5cm and 0.2cm}{[label distance=-1mm]90}{$f_2$}{F2_8}

	\node[rv, draw=newpurple, below left = 0.0cm and 0.5cm of F1_1] (TravelA) {$Travel.alice$};
	\node[rv, draw=newpurple, below right = 0.0cm and 0.5cm of F1_2] (TravelB) {$Travel.bob$};
	\node[rv, draw=newpurple, above right = 0.0cm and 0.5cm of F1_4] (TravelE) {$Travel.eve$};
	\node[rv, draw=newgreen, below left = 0.2cm and -0.8cm of SickA] (TreatAM1) {$Treat.alice.m_1$};
	\node[rv, draw=newgreen, below right = 0.2cm and -0.5cm of TreatAM1] (TreatAM2) {$Treat.alice.m_2$};
	\node[rv, draw=newgreen, below right = 0.2cm and -0.8cm of SickB] (TreatBM1) {$Treat.bob.m_1$};
	\node[rv, draw=newgreen, below left = 0.2cm and -0.5cm of TreatBM1] (TreatBM2) {$Treat.bob.m_2$};
	\node[rv, draw=newgreen, above right = 0.2cm and -0.8cm of SickE] (TreatEM1) {$Treat.eve.m_1$};
	\node[rv, draw=newgreen, above left = 0.2cm and -0.5cm of TreatEM1] (TreatEM2) {$Treat.eve.m_2$};

	\factor{left}{TravelA}{0.2cm}{270}{$f_3$}{F3_1}
	\factor{right}{TravelB}{0.2cm}{270}{$f_3$}{F3_2}
	\factor{right}{TravelE}{0.2cm}{90}{$f_3$}{F3_3}

	\factorcolorshift{newred!80}{F0}{F0_col}{0.2mm}{0mm}
	\factorcolorshift{cborange!80}{F1_1}{F1_1_col}{0.2mm}{0mm}
	\factorcolorshift{cborange!80}{F1_2}{F1_2_col}{0.2mm}{0mm}
	\factorcolorshift{cbbrown!80}{F2_1}{F2_1_col}{0.2mm}{0mm}
	\factorcolorshift{cbbrown!80}{F2_2}{F2_2_col}{0.2mm}{0mm}
	\factorcolorshift{cbbrown!80}{F2_3}{F2_3_col}{0.2mm}{0mm}
	\factorcolorshift{cbbrown!80}{F2_4}{F2_4_col}{0.2mm}{0mm}
	\factorcolorshift{gray!80}{F3_1}{F3_1_col}{0.2mm}{0mm}
	\factorcolorshift{gray!80}{F3_2}{F3_2_col}{0.2mm}{0mm}

	\factorcolorshift{cborange!80}{F1_4}{F1_4_col}{0.2mm}{0mm}
	\factorcolorshift{cbbrown!80}{F2_7}{F2_7_col}{0.2mm}{0mm}
	\factorcolorshift{cbbrown!80}{F2_8}{F2_8_col}{0.2mm}{0mm}
	\factorcolorshift{gray!80}{F3_3}{F3_3_col}{0.2mm}{0mm}

	\begin{pgfonlayer}{bg}
		\draw[newred] (E) -- (F0);
		\draw[newred] (E) -- (F1_1);
		\draw[newred] (E) -- (F2_1);
		\draw[newred] (E) -- (F2_2);
		\draw[newred] (E) -- (F1_2);
		\draw[newred] (E) -- (F2_3);
		\draw[newred] (E) -- (F2_4);
		\draw[newred] (E) -- (F1_4);
		\draw[newred] (E) -- (F2_7);
		\draw[newred] (E) -- (F2_8);
		\draw[newblue] (SickA) -- (F1_1);
		\draw[newblue] (SickA) -- (F2_1);
		\draw[newblue] (SickA) -- (F2_2);
		\draw[newpurple] (TravelA) -- (F1_1);
		\draw[newgreen] (TreatAM1.east) -- (F2_1);
		\draw[newgreen] (TreatAM2) -- (F2_2);
		\draw[newblue] (SickB) -- (F1_2);
		\draw[newblue] (SickB) -- (F2_3);
		\draw[newblue] (SickB) -- (F2_4);
		\draw[newpurple] (TravelB) -- (F1_2);
		\draw[newgreen] (TreatBM1.west) -- (F2_3);
		\draw[newgreen] (TreatBM2) -- (F2_4);
		\draw[newblue] (SickE) -- (F1_4);
		\draw[newblue] (SickE) -- (F2_7);
		\draw[newblue] (SickE) -- (F2_8);
		\draw[newpurple] (TravelE) -- (F1_4);
		\draw[newgreen] (TreatEM1.west) -- (F2_7);
		\draw[newgreen] (TreatEM2) -- (F2_8);
		\draw[newpurple] (TravelA) -- (F3_1);
		\draw[newpurple] (TravelB) -- (F3_2);
		\draw[newpurple] (TravelE) -- (F3_3);
	\end{pgfonlayer}
\end{tikzpicture}}
		\caption{}
		\label{fig:epid_fg_lifagu_04}
	\end{subfigure}
	\caption{An illustration of the steps undertaken by \cref{alg:lifg} (\ac{lifg}) on the input \ac{fg} depicted in \cref{fig:epid_fg_extended_01}: (a) Colouring of known factors (\cref{line:assign_known_colours}), (b) initial colouring of unknown factors (\cref{line:assign_unknown_colours}), (c) refined colouring for unknown factors (\cref{line:reassign_unknown_colours}), and (d) grouping of unknown factors with known factors (\cref{line:assign_unkown_from_known,line:transfer_unkown_from_known}).}
	\label{fig:epid_fg_lifagu}
\end{figure}
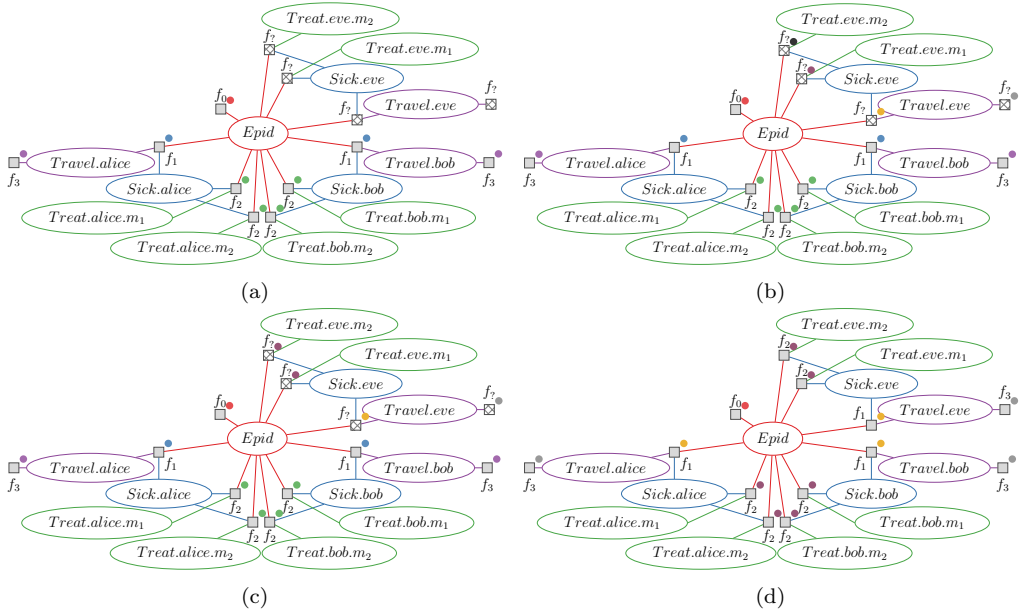

\begin{example}
	Consider again the \ac{fg} $G$ shown in \cref{fig:epid_fg_extended_01} and assume there is no evidence available, i.e., $\boldsymbol E = \emptyset$.
	First, \ac{lifg} assigns colours to all known factors in $G$ according to the potential mappings encoded by their underlying functions.
	In particular, all factors $f_1$ receive the same colour (e.g., blue), all factors $f_2$ receive the same colour (e.g., green), all factors $f_3$ receive the same colour (e.g., purple), and $f_0$ receives a different colour (e.g., red).
	The colourings of the known factors are shown in \cref{fig:epid_fg_lifagu_01}.
	In the next step, all unknown factors $f_{?}$ receive a unique colour and the resulting colourings are given in \cref{fig:epid_fg_lifagu_02}.
	Afterwards, for every unknown factor $f_{?}$, \ac{lifg} searches for factors that are possibly identical to $f_{?}$.
	In this specific example, $f_{?}(Epid, \allowbreak Travel.eve, \allowbreak Sick.eve)$ is possibly identical to the factors $f_1$, $f_{?}(Treat.eve.m_1, \allowbreak Sick.eve, \allowbreak Epid)$ is possibly identical to $f_{?}(Treat.eve.m_2, \allowbreak Sick.eve, \allowbreak Epid)$ as well as to all factors $f_2$, $f_{?}(Treat.eve.m_2, \allowbreak Sick.eve, \allowbreak Epid)$ is possibly identical to $f_{?}(Treat.eve.m_1, \allowbreak Sick.eve, \allowbreak Epid)$ and to all $f_2$, and $f_{?}(Travel.eve)$ is possibly identical to the factors $f_3$.
	Consequently, in \cref{line:reassign_unknown_colours}, \ac{lifg} assigns $f_{?}(Treat.eve.m_1, \allowbreak Sick.eve, \allowbreak Epid)$ and $f_{?}(Treat.eve.m_2, \allowbreak Sick.eve, \allowbreak Epid)$ the same colour (because they are possibly identical and both unknown).
	The new colouring is shown in \cref{fig:epid_fg_lifagu_03}.
	Simultaneously, \ac{lifg} collects for every unknown factor $f_{?}$ a set $C_{f_{?}}$ of possibly identical known factors in \cref{line:add_to_candidates}:
	\begin{itemize}
		\item $C_{f_{?}(Epid, \allowbreak Travel.eve, \allowbreak Sick.eve)} = \{ f_1(Epid, \allowbreak Travel.alice, \allowbreak Sick.alice), \allowbreak f_1(Epid, \allowbreak Travel.bob, \allowbreak Sick.bob) \}$,
		\item $C_{f_{?}(Treat.eve.m_1, \allowbreak Sick.eve, \allowbreak Epid)} = \{ f_2(Treat.alice.m_1, \allowbreak Sick.alice, \allowbreak Epid), \\ \allowbreak f_2(Treat.alice.m_2, \allowbreak Sick.alice, \allowbreak Epid), \allowbreak f_2(Treat.bob.m_1, \allowbreak Sick.bob, \allowbreak Epid), \\ \allowbreak f_2(Treat.bob.m_2, \allowbreak Sick.bob, \allowbreak Epid) \}$,
		\item $C_{f_{?}(Treat.eve.m_2, Sick.eve, Epid)} = \{ f_2(Treat.alice.m_1, \allowbreak Sick.alice, \allowbreak Epid), \\ \allowbreak f_2(Treat.alice.m_2, \allowbreak Sick.alice, \allowbreak Epid), \allowbreak f_2(Treat.bob.m_1, \allowbreak Sick.bob, \allowbreak Epid), \\ \allowbreak f_2(Treat.bob.m_2, \allowbreak Sick.bob, \allowbreak Epid) \}$,
		\item $C_{f_{?}(Travel.eve)} = \{ f_3(Travel.alice), \allowbreak f_3(Travel.bob) \}$.
	\end{itemize}
	Thereafter, \ac{lifg} computes for every candidate set $C_{f_{?}}$ a maximal subset $C_{f_{?}}^{\ell}$ of pairwise possibly identical factors.
	Here, it holds that within each $C_{f_{?}}$, all factors are pairwise possibly identical and hence, we have $C_{f_{?}} = C_{f_{?}}^{\ell}$ for all unknown factors $f_{?}$.
	Due to $C_{f_{?}} = C_{f_{?}}^{\ell}$, we have $\abs{C_{f_{?}}^{\ell}} \mathbin{/} \abs{C_{f_{?}}} = 1$ for all unknown factors $f_{?}$ and thus, the if-condition in \cref{line:if_threshold} is satisfied independent of the choice of $\theta$ in this example.
	In consequence, the factors $f_1$ receive the same colour as $f_{?}(Epid, \allowbreak Travel.eve, \allowbreak Sick.eve)$ and the potentials of the factors $f_1$ are transferred to $f_{?}(Epid, \allowbreak Travel.eve, \allowbreak Sick.eve)$, the factors $f_2$ receive the same colour as $f_{?}(Treat.eve.m_1, \allowbreak Sick.eve, \allowbreak Epid)$ (and $f_{?}(Treat.eve.m_1, \allowbreak Sick.eve, \allowbreak Epid)$) and the potentials of $f_2$ are transferred to $f_{?}(Treat.eve.m_1, \allowbreak Sick.eve, \allowbreak Epid)$ and $f_{?}(Treat.eve.m_1, \allowbreak Sick.eve, \allowbreak Epid)$, and the factors $f_3$ receive the same colour as $f_{?}(Travel.eve)$ and the potentials of $f_3$ are transferred to $f_{?}(Travel.eve)$.
	The resulting colourings are shown in \cref{fig:epid_fg_lifagu_04}, where all unknown factors $f_{?}$ have been replaced by known factors.
	Finally, \ac{lifg} calls \ac{acp} on the resulting graph $G$ from \cref{fig:epid_fg_lifagu_04} to obtain the lifted representation of $G$ illustrated in \cref{fig:epid_pm}.
\end{example}

The purpose of the threshold $\theta$ is to control the required agreement of known factors before grouping unknown factors with known factors as it is possible for an unknown factor to be possibly identical to multiple known factors having different potentials.
A larger value for $\theta$ requires a higher agreement, e.g., $\theta = 1$ requires all candidates to have identical potentials.
Note that all known factors in $C_{f_i}^{\ell}$ are guaranteed to have identical potentials (because otherwise they would violate \cref{itm:lifagu_possibly_identical_itm_2} of \cref{def:lifagu_possibly_identical} and hence not be pairwise possibly identical) and thus, their potentials can be transferred to $f_i$.
Consequently, the output of \ac{lifg} is guaranteed to contain only known factors and hence, \ac{lifg} ensures a well-defined semantics if $C_{f_i}^{\ell}$ is non-empty for each unknown factor $f_i$ and the threshold is sufficiently small (e.g., zero) to group each unknown factor with at least one known factor.\footnote{As the semantics of an \ac{fg} is given by a product over its factors, the semantics of the \ac{fg} is only well-defined if all input-output mappings of the factors are known.}

\begin{theorem}
	Given that for every unknown factor $f_i$ there is at least one known factor that is possibly identical to $f_i$ in an \ac{fg} $G$, \ac{lifg} is able to replace all unknown potentials in $G$ by known potentials.
\end{theorem}
\begin{proof}
	Let $G$ be an \ac{fg} with known factors $\boldsymbol{F} = \{f_1, \dots, f_m\}$ and unknown factors $\boldsymbol{F'} = \{f'_1, \dots, f'_z\}$ such that for each unknown factor $f_i \in \boldsymbol{F'}$ there exists at least one known factor $f_j \in \boldsymbol{F}$ such that $f_i \approx f_j$.
	Then, it is guaranteed for each unknown factor $f_i$ that \cref{line:add_to_candidates} in \cref{alg:lifg} is executed at least once and thus $C_{f_i}$ is non-empty for all unknown factors $f_i$.
	Afterwards, as $C_{f_i} \neq \emptyset$ holds for all unknown factors $f_i$, it holds that in \cref{line:max_subset_candidates} there is at least one element in $C_{f_i}^{\ell}$ for every unknown factor $f_i$.
	Therefore, if we set $\theta = 0$, the if-condition in \cref{line:if_threshold} passes successfully for each unknown factor $f_i$, resulting both in $f_i$ being assigned the same colour as at least one known factor $f_j \in C_{f_i}^{\ell}$ as well as in the transfer of $f_j$'s potentials to $f_i$.
\end{proof}

The threshold $\theta$ defines the trade-off between transferring as much potentials from known factors to unknown factors as possible and avoiding incorrect groupings of unknown factors.
While a small threshold $\theta$ might be able to provide guarantees about a well-defined semantics of the output of \ac{lifg}, a larger threshold $\theta$ might be able to reduce incorrectly grouped unknown factors.
In particular, for an unknown factor $f_i$, it is generally possible that $C_{f_i}^{\ell}$ is not unique, i.e., there are multiple maximal subsets of candidates of known factors $f_i$ might be grouped with.
The threshold $\theta$ can be used to avoid such scenarios by setting $\theta > 0.5$.

\begin{theorem}
	Let $C_{f_i}$ be the set of known factors possibly identical to an unknown factor $f_i$ and $C_{f_i}^{\ell}$ a maximal subset of $C_{f_i}$ with  $f_j \approx f_k$ for all $f_j$, $f_k \in C_{f_i}^{\ell}$.
	Then, $C_{f_i}^{\ell}$ is guaranteed to be unique if $\abs{C_{f_i}^{\ell}} \mathbin{/} \abs{C_{f_i}} > 0.5$.
\end{theorem}
\begin{proof}
	Let $C_{f_i}$ be a set of known factors possibly identical to an unknown factor $f_i$, $C_{f_i}^{\ell} \subseteq C_{f_i}$ a maximal subset such that $f_j \approx f_k$ holds for all $f_j, f_k \in C_{f_i}^{\ell}$, and $\abs{C_{f_i}^{\ell}} \mathbin{/} \abs{C_{f_i}} > 0.5$.
	For the sake of contradiction, assume that there is another maximal subset $C_{f_i}^{\ell'} \subseteq C_{f_i}$ containing only pairwise possibly identical factors with $C_{f_i}^{\ell'} \neq C_{f_i}^{\ell}$ and $\abs{C_{f_i}^{\ell'}} = \abs{C_{f_i}^{\ell}}$.
	As $\abs{C_{f_i}^{\ell}} > 0.5 \cdot \abs{C_{f_i}}$ holds, there must be a factor, say $f_j$, with $f_j \in C_{f_i}^{\ell'} \cap C_{f_i}^{\ell}$.
	Consequently, $f_j$ is pairwise possibly identical to all $f_k \in C_{f_i}^{\ell'}$ and to all $f_l \in C_{f_i}^{\ell}$, and as all $f_k$ and $f_l$ are known, this implies that both all $f_k$ as well as all $f_l$ have the same potentials as $f_j$, meaning the $f_k$ and $f_l$ must be pairwise possibly identical as well.
	This implies that $C_{f_i}^{\ell'} = C_{f_i}^{\ell}$ because if there were a factor $f_r$ with $f_r \in C_{f_i}^{\ell'}$ and $f_r \notin C_{f_i}^{\ell}$, then $C_{f_i}^{\ell}$ can not be maximal as $f_r$ is pairwise possibly identical to all $f_l \in C_{f_i}^{\ell}$.
	A contradiction to our assumption that $C_{f_i}^{\ell'} \neq C_{f_i}^{\ell}$.
\end{proof}

To close this section, we prove that \ac{lifg} is a generalisation of \ac{acp}, i.e., both algorithms compute the same result for input \acp{fg} containing only known factors (independent of $\theta$ because $\theta$ only affects unknown factors).

\begin{theorem}
	Given an \ac{fg} that contains only known factors, \ac{acp} and \ac{lifg} output identical groupings of \acp{rv} and factors, respectively.
\end{theorem}
\begin{proof}
	Let $G$ be an \ac{fg} containing only known factors.
	Then, only \cref{line:assign_known_colours,line:call_cp} of \cref{alg:lifg} are executed---which is equivalent to calling \ac{acp} on $G$.
\end{proof}

Before we evaluate the practical performance of \ac{lifg} empirically, we extend \ac{lifg} to incorporate background knowledge about factors belonging to the same individual object in the upcoming section.

\section{Incorporating Background Knowledge in \acs{lifg}} \label{sec:lifg_bk}
We start this section by first defining the concept of background knowledge and afterwards elaborate on how given background knowledge can be incorporated into \ac{lifg}.
Informally speaking, in our setting, background knowledge specifies which factors belong to the same individual object.

\begin{definition}[Background Knowledge]
	Let $G$ denote an \ac{fg} with known factors $\boldsymbol{F} = \{f_1, \dots, f_m\}$ and unknown factors $\boldsymbol{F'} = \{f'_1, \dots, f'_z\}$.
	Then, \emph{background knowledge} $\mathcal{K} = \langle \boldsymbol{K}_1, \dots, \boldsymbol{K}_d \rangle$ is a collection of sets of factors $\boldsymbol{K}_1, \dots, \boldsymbol{K}_d$ such that $\boldsymbol{K}_i \subseteq \boldsymbol F \cup \boldsymbol F'$, $i \in \{1, \dots, d\}$, specifies a set of factors belonging to the same individual $i$.
	We say that $\mathcal{K}$ is valid if $\boldsymbol{K}_i \cap \boldsymbol{K}_j = \emptyset$ for all pairs of $\boldsymbol{K}_i \in \mathcal{K}$ and $\boldsymbol{K}_j \in \mathcal{K}$ with $i \neq j$.
\end{definition}

As background knowledge tells us which factors belong to the same individual, we are only interested in valid background knowledge, i.e., background knowledge in which each factor belongs to at most one individual.
From now on, we therefore use the term \emph{background knowledge} to refer to valid background knowledge only.
Note that background knowledge might not be available for every individual and it is also possible that there is no background knowledge available at all.
In general, there are (at least) two possible approaches to make use of background knowledge in \ac{lifg}:
\begin{enumerate}
	\item Group unknown factors with known factors solely based on the available background knowledge instead of searching for a maximal subset of known factor candidates which is then used to group unknown factors with known factors if a given threshold is reached.
	\item Make the decision of whether an unknown factor should be grouped with a set of known factors based on a combination of the threshold and background knowledge.
\end{enumerate}
Using only the available background knowledge to group unknown factors with known factors is mostly not desirable because background knowledge often is limited.
Combining the threshold with background knowledge, however, possibly reduces the ambiguity for possible transfers of potentials from known factors to unknown factors.
We next explain how the threshold can be combined with background knowledge in \ac{lifg} to determine which set of known factors an unknown factor should be grouped with.

\begin{figure}[t]
	\centering
	\resizebox{\textwidth}{!}{\begin{tikzpicture}[rv/.append style={fill=white}]
	\node[rv, draw=newred] (E) {$Epid$};

	\factor{above left}{E}{0.2cm and 0.3cm}{[label distance=-1mm]90}{$f_0$}{F0}

	\factor{below left}{E}{-0.1cm and 2cm}{[label distance=-2mm]315}{$f_1$}{F1_1}
	\factor{below right}{E}{-0.1cm and 2cm}{[label distance=-2mm]215}{$f_1$}{F1_2}
	\factor{above left}{E}{-0.1cm and 2cm}{[label distance=-2mm]45}{$f'_1$}{F1_3}
	\factor{above right}{E}{-0.1cm and 2cm}{[label distance=-2mm]135}{$f'_1$}{F1_4}

	\node[rv, draw=newblue, below = 0.6cm of F1_1] (SickA) {$Sick.alice$};
	\node[rv, draw=newblue, below = 0.6cm of F1_2] (SickB) {$Sick.bob$};
	\node[rv, draw=newblue, above = 0.6cm of F1_3] (SickD) {$Sick.dave$};
	\node[rv, draw=newblue, above = 0.6cm of F1_4] (SickE) {$Sick.eve$};

	\factor{right}{SickA}{0.5cm}{[label distance=-1mm]270}{$f_2$}{F2_1}
	\factor{below right}{F2_1}{0.5cm and 0.2cm}{[label distance=-1mm]270}{$f_2$}{F2_2}

	\factor{left}{SickB}{0.5cm}{[label distance=-1mm]270}{$f_2$}{F2_3}
	\factor{below left}{F2_3}{0.5cm and 0.2cm}{[label distance=-1mm]270}{$f_2$}{F2_4}

	\factor{right}{SickD}{0.5cm}{[label distance=-1mm]90}{$f'_2$}{F2_5}
	\factor{above right}{F2_5}{0.5cm and 0.2cm}{[label distance=-1mm]90}{$f'_2$}{F2_6}

	\ufactor{left}{SickE}{0.5cm}{[label distance=-1mm]90}{$f_{?}$}{F2_7}
	\ufactor{above left}{F2_7}{0.5cm and 0.2cm}{[label distance=-1mm]90}{$f_{?}$}{F2_8}

	\node[rv, draw=newpurple, below left = 0.0cm and 0.5cm of F1_1] (TravelA) {$Travel.alice$};
	\node[rv, draw=newpurple, below right = 0.0cm and 0.5cm of F1_2] (TravelB) {$Travel.bob$};
	\node[rv, draw=newpurple, above left = 0.0cm and 0.5cm of F1_3] (TravelD) {$Travel.dave$};
	\node[rv, draw=newpurple, above right = 0.0cm and 0.5cm of F1_4] (TravelE) {$Travel.eve$};
	\node[rv, draw=newgreen, below left = 0.2cm and -0.8cm of SickA] (TreatAM1) {$Treat.alice.m_1$};
	\node[rv, draw=newgreen, below right = 0.2cm and -0.5cm of TreatAM1] (TreatAM2) {$Treat.alice.m_2$};
	\node[rv, draw=newgreen, below right = 0.2cm and -0.8cm of SickB] (TreatBM1) {$Treat.bob.m_1$};
	\node[rv, draw=newgreen, below left = 0.2cm and -0.5cm of TreatBM1] (TreatBM2) {$Treat.bob.m_2$};
	\node[rv, draw=newgreen, above left = 0.2cm and -0.8cm of SickD] (TreatDM1) {$Treat.dave.m_1$};
	\node[rv, draw=newgreen, above right = 0.2cm and -0.5cm of TreatDM1] (TreatDM2) {$Treat.dave.m_2$};
	\node[rv, draw=newgreen, above right = 0.2cm and -0.8cm of SickE] (TreatEM1) {$Treat.eve.m_1$};
	\node[rv, draw=newgreen, above left = 0.2cm and -0.5cm of TreatEM1] (TreatEM2) {$Treat.eve.m_2$};

	\factor{left}{TravelA}{0.2cm}{270}{$f_3$}{F3_1}
	\factor{right}{TravelB}{0.2cm}{270}{$f_3$}{F3_2}
	\factor{left}{TravelD}{0.2cm}{90}{$f'_3$}{F3_3}
	\factor{right}{TravelE}{0.2cm}{90}{$f'_3$}{F3_4}

	\begin{pgfonlayer}{bg}
		\draw[newred] (E) -- (F0);
		\draw[newred] (E) -- (F1_1);
		\draw[newred] (E) -- (F2_1);
		\draw[newred] (E) -- (F2_2);
		\draw[newred] (E) -- (F1_2);
		\draw[newred] (E) -- (F2_3);
		\draw[newred] (E) -- (F1_3);
		\draw[newred] (E) -- (F2_4);
		\draw[newred] (E) -- (F1_4);
		\draw[newred] (E) -- (F2_5);
		\draw[newred] (E) -- (F2_6);
		\draw[newred] (E) -- (F2_7);
		\draw[newred] (E) -- (F2_8);
		\draw[newblue] (SickA) -- (F1_1);
		\draw[newblue] (SickA) -- (F2_1);
		\draw[newblue] (SickA) -- (F2_2);
		\draw[newpurple] (TravelA) -- (F1_1);
		\draw[newgreen] (TreatAM1.east) -- (F2_1);
		\draw[newgreen] (TreatAM2) -- (F2_2);
		\draw[newblue] (SickB) -- (F1_2);
		\draw[newblue] (SickB) -- (F2_3);
		\draw[newblue] (SickB) -- (F2_4);
		\draw[newpurple] (TravelB) -- (F1_2);
		\draw[newgreen] (TreatBM1.west) -- (F2_3);
		\draw[newgreen] (TreatBM2) -- (F2_4);
		\draw[newblue] (SickD) -- (F1_3);
		\draw[newblue] (SickD) -- (F2_5);
		\draw[newblue] (SickD) -- (F2_6);
		\draw[newpurple] (TravelD) -- (F1_3);
		\draw[newgreen] (TreatDM1.east) -- (F2_5);
		\draw[newgreen] (TreatDM2) -- (F2_6);
		\draw[newblue] (SickE) -- (F1_4);
		\draw[newblue] (SickE) -- (F2_7);
		\draw[newblue] (SickE) -- (F2_8);
		\draw[newpurple] (TravelE) -- (F1_4);
		\draw[newgreen] (TreatEM1.west) -- (F2_7);
		\draw[newgreen] (TreatEM2) -- (F2_8);
		\draw[newpurple] (TravelA) -- (F3_1);
		\draw[newpurple] (TravelB) -- (F3_2);
		\draw[newpurple] (TravelD) -- (F3_3);
		\draw[newpurple] (TravelE) -- (F3_4);
	\end{pgfonlayer}
\end{tikzpicture}}
	\caption{A slightly modified and extended version of the epidemic example depicted in \cref{fig:epid_fg_extended_01}. The factors $f_{?}$ are unknown and the input-output pairs of the remaining factors are again omitted for brevity. Note that the factors $f_1$ encode a different underlying function than the factors $f'_1$ and the factors $f_2$ encode a different underlying function than $f'_2$.}
	\label{fig:epid_fg_extended_02}
\end{figure}

Let us take a look at the \ac{fg} $G$ depicted in \cref{fig:epid_fg_extended_02}, where four individuals $alice$, $bob$, $dave$, and $eve$ are part of the model.
In this example, $alice$ and $bob$ belong to the same group of identically behaving individuals as they share the same potentials for the factors $f_1$, $f_2$, and $f_3$, i.e., it holds that $f_1(Travel.alice, Sick.alice, Epid) = f_1(Travel.bob, Sick.bob, Epid)$ for all possible assignments of the arguments of the factors $f_1$, and analogously for all factors $f_2$ and $f_3$.
Further, let us assume that $f'_1 \neq f_1$, $f'_2 \neq f_2$, and $f'_3 \neq f_3$ hold---that is, $dave$ belongs to a different group than $alice$ and $bob$ (formally, this can be encoded in a \ac{pfg} by using constraints).
Additionally, $G$ contains another individual $eve$, for which we have only limited data available.
In particular, we do not know the exact potentials for the factors $f_{?}(Treat.eve.m_1, Sick.eve, Epid)$ and $f_{?}(Treat.eve.m_2, Sick.eve, Epid)$.
However, we do know the potentials for $f'_1(Epid, Travel.eve, Sick.eve)$ and for $f'_3(Travel.eve)$, and for the sake of the example, let us assume that we are also given the background knowledge $\mathcal K = \langle \boldsymbol{K}_1, \boldsymbol{K}_2 \rangle$ with
\begin{itemize}
	\item $\boldsymbol{K}_1 \mkern-3.9mu = \mkern-3.9mu \{ f'_1(Epid, \allowbreak \mkern-1.4mu Travel.eve, \allowbreak \mkern-1.4mu Sick.eve), \allowbreak \mkern-1.4mu f_{?}(Treat.eve.m_1, \allowbreak \mkern-1.4mu Sick.eve, \allowbreak \mkern-1.4mu Epid), \allowbreak \mkern-1.4mu f_{?}(Treat.eve.m_2, \allowbreak \mkern-1.4mu Sick.eve, \allowbreak \mkern-1.4mu Epid), \allowbreak \mkern-1.4mu f'_3(Travel.eve) \}$ and
	\item $\boldsymbol{K}_2 \mkern-3.9mu = \mkern-3.9mu \{ f'_1(Epid, \allowbreak Travel.dave, \allowbreak Sick.dave), \allowbreak f'_2(Treat.dave.m_1, \allowbreak Sick.dave, \allowbreak Epid), \allowbreak f'_2(Treat.dave.m_2, \allowbreak Sick.dave, \allowbreak Epid), \allowbreak f'_3(Travel.dave) \}$.
\end{itemize}
In other words, we know which factors belong to the individuals $dave$ and $eve$ but we do not have any information about the remaining factors in $G$.

Since it holds that $dave$ and $eve$ share the same potentials for the factors $f'_1$, i.e., $f'_1(Travel.dave, Sick.dave, Epid)$ = $f'_1(Travel.eve, Sick.eve, Epid)$ for all possible assignments of the arguments of the factors $f'_1$, we know that with respect to $f'_1$, $dave$ and $eve$ belong to a group of identically behaving individuals (analogously for $f'_3$).
However, we do not know whether the factors $f_{?}$ should be grouped with the factors $f_2$, $f'_2$, or none of them.
At this point, we can apply our background knowledge $\mathcal K$: As $dave$ and $eve$ share the same potentials for the factors $f'_1$ as well as for $f'_3$ and we know that the factors $f_{?}$ belong to $eve$ as well as that the factors $f'_2$ belong to $dave$, we might want to decide to group the factors $f_{?}$ with the factors $f'_2$ to achieve that $dave$ and $eve$ are grouped together.

Generally, we thus aim to prefer grouping an unknown factor with a group of known factors that is supported by the available background knowledge.
The next definition formalises the idea of supporting background knowledge.

\begin{definition}[Supporting Background Knowledge] \label{def:supporting_bk}
	Let $\mathcal{K} \mkern-1.8mu = \mkern-1.8mu \langle \boldsymbol{K}_1, \dots, \boldsymbol{K}_d \rangle$ be given background knowledge, $C_{f_i}$ a set of known factors possibly identical to an unknown factor $f_i$, and $C_{f_i}^{s}$ a subset of $C_{f_i}$ such that $f_j \approx f_k$ holds for all $f_j, f_k \in C_{f_i}^{s}$.
	We say that $C_{f_i}^{s}$ is \emph{supported} by $\mathcal{K}$ if
	\begin{enumerate}
		\item there exists no set $\boldsymbol{K}_i \in \mathcal{K}$ such that $f_i \in \boldsymbol{K}_i$, or
		\item there exists a set $\boldsymbol{K}_i \in \mathcal{K}$ such that $f_i \in \boldsymbol{K}_i$, and
			\begin{enumerate}[i.]
				\item for all known factors $f_{\ell} \in \boldsymbol{K}_i$ it holds that there exists at most one set $\boldsymbol{K}_o \in \mathcal{K}$, $\boldsymbol{K}_o \neq \boldsymbol{K}_i$, which contains a factor having the same colour as $f_{\ell}$ and all factors in $C_{f_i}^{s}$ have the same colour as $f_{\ell}$, and
				\item the set $\boldsymbol{K}_o$ is the same set for all known factors $f_{\ell} \in \boldsymbol{K}_i$.
			\end{enumerate}
	\end{enumerate}
\end{definition}

The notion of supporting background knowledge can be integrated into \ac{lifg} by searching for subsets of candidates of known factors that are pairwise possibly identical and that are supported by the given background knowledge instead of searching for a maximal subset of candidates.
More specifically, in \cref{line:max_subset_candidates} in \cref{alg:lifg}, \ac{lifg} now computes all subsets $C_{f_i}^{s} \subseteq C_{f_i}$ such that $f_j \approx f_k$ holds for all $f_j, f_k \in C_{f_i}^{s}$ and then checks for all subsets $C_{f_i}^{s}$ whether they are supported by the given background knowledge.
If no subset is supported by the given background knowledge, \ac{lifg} proceeds as before and takes the maximal subset $C_{f_i}^{\ell} \subseteq C_{f_i}$ for the transfer of potentials from known factors to unknown factors.
Otherwise (i.e., in case at least one of the subsets $C_{f_i}^{s}$ is supported by the given background knowledge), \ac{lifg} takes the maximal subset of all subsets that are supported by the given background knowledge for the transfer of potentials from known factors to unknown factors.
The idea behind this approach is that \ac{lifg} proceeds as usual if background knowledge is either missing or does not uniquely hint at a specific individual whose known factors should be used for grouping.
In cases where the known factors of the individual to which $f_i$ belongs might be grouped with known factors from various other individuals, we do not know which of these individuals should be chosen for grouping.
Thus, we require that the known factors of the individual to which $f_i$ belongs might be grouped with the known factors of a unique other individual to make use of given background knowledge.

Note that a situation like the one we considered in our toy example from \cref{fig:epid_fg_extended_02} is abundant in many real-world applications.
For example, when a new patient arrives at a hospital, there is limited data available as not all measurements are taken immediately, i.e., it is conceivable that a first examination determines the current blood pressure of the patient while measurements for other attributes are not conducted yet.
Background knowledge in combination with partial measurements can help assigning the new patient to a group of indistinguishable patients, thereby allowing to draw tentative conclusions about which measurement to take next or which treatment to apply.

So far, we have introduced \ac{lifg} and analysed its theoretical properties.
We have also extended \ac{lifg} to incorporate background knowledge.
Next, we investigate the practical performance of \ac{lifg} empirically.

\section{Empirical Evaluation} \label{sec:eval}
In this section, we present the results of the empirical evaluation for \ac{lifg}.
To evaluate the performance of \ac{lifg}, we start with a non-parameterised \ac{fg} $G$ where all factors are known, serving as our ground truth.
Afterwards, we remove the potential mappings for $5$ to $20$ percent of the factors in $G$, yielding an incomplete \ac{fg} $G'$ on which \ac{lifg} is run to obtain a \ac{pfg} $G_{\textsc{\ac{lifg}}}$.
Each factor $f'$ whose potentials are removed is chosen randomly under the constraint that there exists at least one other factor with known potentials that is possibly identical to $f'$.
This constraint corresponds to the assumption that there exists at least one group to which a new individual can be added and it ensures that after running \ac{lifg}, probabilistic inference can be performed for evaluation purposes.
Clearly, in our evaluation setting, there is not only a single new individual but instead a set of new individuals, given by the set of factors whose potentials are missing.
There is no background knowledge available in our experiments.
We use a parameter $d=2,4,8,16,32,64,128,256$ to control the size of the \ac{fg} $G$ (and thus, the size of $G'$).
More precisely, for each choice of $d$, we evaluate multiple graph structures for input \acp{fg}, which contain between $2d$ and $3d$ \acp{rv} (and factors, respectively).
The potentials of the factors are randomly generated such that the ground truth $G$ contains between three and five (randomly chosen) cohorts of \acp{rv} which behave identically and thus should be grouped together.
We evaluate different choices for the sizes of the cohorts: There is one cohort which contains a proportion of $p \in \{0.2, 0.3, 0.5, 0.7, 0.9\}$ of all \acp{rv} in $G$ whereas the other cohorts share the remaining proportion of $1-p$ of the \acp{rv} from $G$ uniformly at random.

We set $\theta = 0$ to ensure that each unknown factor is grouped with at least one known factor to be able to perform lifted probabilistic inference on $G_{\textsc{\ac{lifg}}}$ for evaluation.
To assess the error made by \ac{lifg} for each choice of $d$, we pose between three and four different queries to the ground truth $G$ and to $G_{\textsc{\ac{lifg}}}$, respectively.
For each query, we compute the \ac{kld}~\citep{Kullback1951a} between the resulting probability distributions for the ground truth $G$ and $G_{\textsc{\ac{lifg}}}$ to measure the similarity of the query results.
The \ac{kld} measures the difference between two distributions $P$ and $Q$ and is defined as
\begin{align}
	\KLD(P \parallel Q) &= \sum\limits_{x} P(x) \cdot \log \left( \frac{P(x)}{Q(x)} \right).
\end{align}
If the distributions $P$ and $Q$ are identical, the \ac{kld} is zero and the larger the \ac{kld}, the more $P$ and $Q$ differ from each other.

\begin{figure}[t]
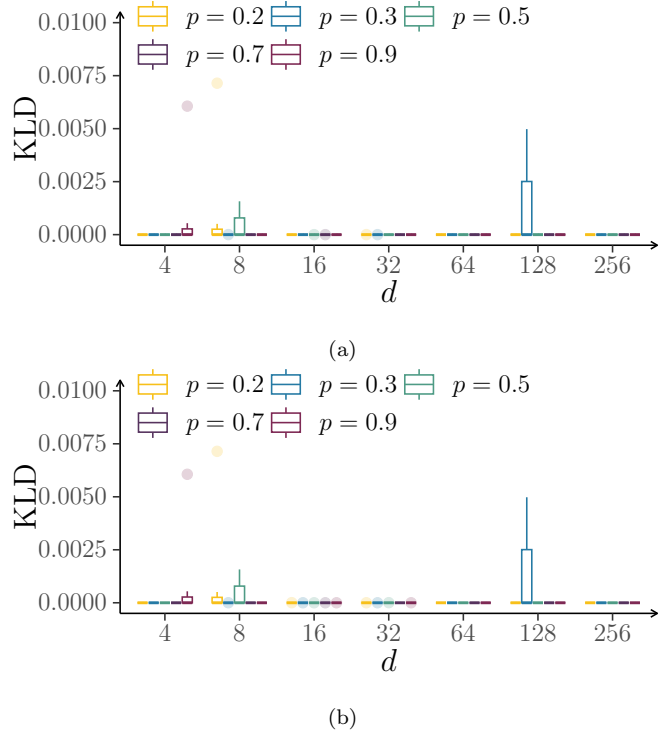

	\centering
	\begin{subfigure}{\linewidth}
		\centering
		\input{files/lifagu-plot-kldiv-p=0.05.tex}
		\caption{}
		\label{fig:lifagu-plot-kldiv-p=0.05}
	\end{subfigure}

	\begin{subfigure}{\linewidth}
		\centering
		\input{files/lifagu-plot-kldiv-p=0.1.tex}
		\caption{}
		\label{fig:lifagu-plot-kldiv-p=0.1}
	\end{subfigure}
	\caption{(a) A boxplot showing the measured \acp{kld} for input \acp{fg} where roughly $5$ percent of the factors are unknown, and (b) a boxplot showing the measured \ac{kld} for input \acp{fg} where roughly $10$ percent of the factors are unknown.}
	\label{fig:lifagu-plot-kldiv-p=0.05-0.1}
\end{figure}
\begin{figure}[t]
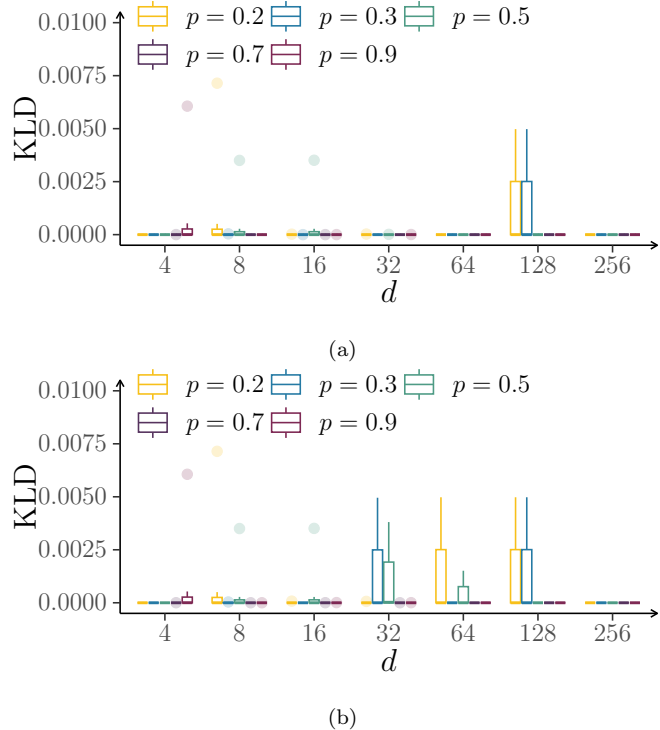

	\centering
	\begin{subfigure}{\linewidth}
		\centering
		\input{files/lifagu-plot-kldiv-p=0.15.tex}
		\caption{}
		\label{fig:lifagu-plot-kldiv-p=0.15}
	\end{subfigure}

	\begin{subfigure}{\linewidth}
		\centering
		\input{files/lifagu-plot-kldiv-p=0.2.tex}
		\caption{}
		\label{fig:lifagu-plot-kldiv-p=0.2}
	\end{subfigure}
	\caption{(a) A boxplot showing the measured \acp{kld} for input \acp{fg} where roughly $15$ percent of the factors are unknown, and (b) a boxplot showing the measured \ac{kld} for input \acp{fg} where roughly $20$ percent of the factors are unknown.}
	\label{fig:lifagu-plot-kldiv-p=0.15-0.2}
\end{figure}

In \cref{fig:lifagu-plot-kldiv-p=0.05-0.1,fig:lifagu-plot-kldiv-p=0.15-0.2}, we present boxplots showing the distributions of the measured \acp{kld} for the different choices of $d$ and $p$.
We can observe that in every scenario, the \ac{kld} is close to zero, indicating that the query results for $G_{\textsc{\ac{lifg}}}$ are close to the query results for the ground truth $G$ in practice.
Interestingly, there are no major differences between different choices of the parameters $d$ and $p$.
Even though there are some choices of $d$ and $p$ having slightly larger \acp{kld} than other choices for $d$ and $p$, there are especially no systematic differences between the distribution of the cohort sizes.
Note that even the largest values for the \ac{kld} are still below $0.01$ here.

Given our assumptions, a new individual actually belongs to a cohort and most cohorts behave not completely different.
So normally, we trade off accuracy of query results for the ability to perform inference, which otherwise would not be possible at all.
If the semantics of the model cannot be fixed, missing potentials need to be guessed to be able to perform inference at all, probably resulting in worse errors.
As we basically perform unsupervised clustering, errors might happen whenever unknown factors are grouped with known factors.
The error might be further reduced by increasing the effort when searching for known factors that are possible candidates for grouping with an unknown factor---for example, it is conceivable to increase the size of the neighbourhood during the search for possible identical factors at the expense of a higher run time expenditure for \ac{lifg}.

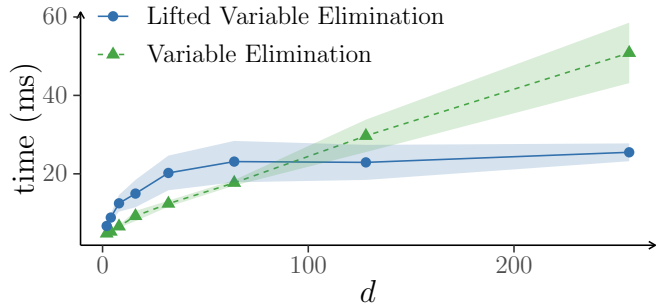
\begin{figure}[t]
	\centering
\begin{tikzpicture}[x=1pt,y=1pt]
\definecolor{fillColor}{RGB}{255,255,255}
\path[use as bounding box,fill=fillColor,fill opacity=0.00] (0,0) rectangle (252.94,122.86);
\begin{scope}
\path[clip] (  0.00,  0.00) rectangle (252.94,122.86);
\definecolor{drawColor}{RGB}{255,255,255}
\definecolor{fillColor}{RGB}{255,255,255}

\path[draw=drawColor,line width= 0.6pt,line join=round,line cap=round,fill=fillColor] (  0.00,  0.00) rectangle (252.94,122.86);
\end{scope}
\begin{scope}
\path[clip] ( 30.03, 29.10) rectangle (247.44,117.36);
\definecolor{fillColor}{RGB}{255,255,255}

\path[fill=fillColor] ( 30.03, 29.10) rectangle (247.44,117.36);
\definecolor{drawColor}{RGB}{50,113,173}

\path[draw=drawColor,line width= 0.6pt,line join=round] ( 39.91, 36.39) --
	( 41.47, 39.57) --
	( 44.58, 44.97) --
	( 50.81, 48.65) --
	( 63.26, 56.46) --
	( 88.16, 60.76) --
	(137.96, 60.46) --
	(237.56, 64.28);
\definecolor{drawColor}{RGB}{70,165,69}

\path[draw=drawColor,line width= 0.6pt,dash pattern=on 2pt off 2pt ,line join=round] ( 39.91, 33.54) --
	( 41.47, 34.25) --
	( 44.58, 36.21) --
	( 50.81, 40.16) --
	( 63.26, 44.83) --
	( 88.16, 52.73) --
	(137.96, 70.50) --
	(237.56,101.91);
\definecolor{fillColor}{RGB}{70,165,69}

\path[fill=fillColor] ( 63.26, 47.89) --
	( 65.90, 43.31) --
	( 60.61, 43.31) --
	cycle;

\path[fill=fillColor] ( 41.47, 37.30) --
	( 44.11, 32.72) --
	( 38.83, 32.72) --
	cycle;

\path[fill=fillColor] ( 88.16, 55.78) --
	( 90.80, 51.20) --
	( 85.51, 51.20) --
	cycle;

\path[fill=fillColor] ( 39.91, 36.59) --
	( 42.55, 32.01) --
	( 37.27, 32.01) --
	cycle;

\path[fill=fillColor] (237.56,104.96) --
	(240.21,100.38) --
	(234.92,100.38) --
	cycle;

\path[fill=fillColor] (137.96, 73.55) --
	(140.60, 68.97) --
	(135.32, 68.97) --
	cycle;

\path[fill=fillColor] ( 50.81, 43.21) --
	( 53.45, 38.63) --
	( 48.16, 38.63) --
	cycle;

\path[fill=fillColor] ( 44.58, 39.26) --
	( 47.22, 34.69) --
	( 41.94, 34.69) --
	cycle;
\definecolor{fillColor}{RGB}{50,113,173}

\path[fill=fillColor] ( 63.26, 56.46) circle (  1.96);

\path[fill=fillColor] ( 41.47, 39.57) circle (  1.96);

\path[fill=fillColor] ( 88.16, 60.76) circle (  1.96);

\path[fill=fillColor] ( 39.91, 36.39) circle (  1.96);

\path[fill=fillColor] (237.56, 64.28) circle (  1.96);

\path[fill=fillColor] (137.96, 60.46) circle (  1.96);

\path[fill=fillColor] ( 50.81, 48.65) circle (  1.96);

\path[fill=fillColor] ( 44.58, 44.97) circle (  1.96);
\definecolor{fillColor}{RGB}{50,113,173}

\path[fill=fillColor,fill opacity=0.20] ( 39.91, 37.18) --
	( 41.47, 40.96) --
	( 44.58, 48.14) --
	( 50.81, 53.83) --
	( 63.26, 63.03) --
	( 88.16, 68.59) --
	(137.96, 67.09) --
	(237.56, 67.65) --
	(237.56, 60.90) --
	(137.96, 53.82) --
	( 88.16, 52.93) --
	( 63.26, 49.90) --
	( 50.81, 43.47) --
	( 44.58, 41.80) --
	( 41.47, 38.19) --
	( 39.91, 35.60) --
	cycle;

\path[] ( 39.91, 37.18) --
	( 41.47, 40.96) --
	( 44.58, 48.14) --
	( 50.81, 53.83) --
	( 63.26, 63.03) --
	( 88.16, 68.59) --
	(137.96, 67.09) --
	(237.56, 67.65);

\path[] (237.56, 60.90) --
	(137.96, 53.82) --
	( 88.16, 52.93) --
	( 63.26, 49.90) --
	( 50.81, 43.47) --
	( 44.58, 41.80) --
	( 41.47, 38.19) --
	( 39.91, 35.60);
\definecolor{fillColor}{RGB}{70,165,69}

\path[fill=fillColor,fill opacity=0.20] ( 39.91, 33.96) --
	( 41.47, 34.66) --
	( 44.58, 36.66) --
	( 50.81, 42.10) --
	( 63.26, 45.98) --
	( 88.16, 53.78) --
	(137.96, 76.64) --
	(237.56,113.35) --
	(237.56, 90.46) --
	(137.96, 64.36) --
	( 88.16, 51.68) --
	( 63.26, 43.69) --
	( 50.81, 38.22) --
	( 44.58, 35.76) --
	( 41.47, 33.83) --
	( 39.91, 33.11) --
	cycle;

\path[] ( 39.91, 33.96) --
	( 41.47, 34.66) --
	( 44.58, 36.66) --
	( 50.81, 42.10) --
	( 63.26, 45.98) --
	( 88.16, 53.78) --
	(137.96, 76.64) --
	(237.56,113.35);

\path[] (237.56, 90.46) --
	(137.96, 64.36) --
	( 88.16, 51.68) --
	( 63.26, 43.69) --
	( 50.81, 38.22) --
	( 44.58, 35.76) --
	( 41.47, 33.83) --
	( 39.91, 33.11);
\end{scope}
\begin{scope}
\path[clip] (  0.00,  0.00) rectangle (252.94,122.86);
\definecolor{drawColor}{RGB}{0,0,0}

\path[draw=drawColor,line width= 0.6pt,line join=round] ( 30.03, 29.10) --
	( 30.03,117.36);

\path[draw=drawColor,line width= 0.6pt,line join=round] ( 31.45,114.89) --
	( 30.03,117.36) --
	( 28.61,114.89);
\end{scope}
\begin{scope}
\path[clip] (  0.00,  0.00) rectangle (252.94,122.86);
\definecolor{drawColor}{gray}{0.30}

\node[text=drawColor,anchor=base east,inner sep=0pt, outer sep=0pt, scale=  0.80] at ( 25.08, 53.37) {20};

\node[text=drawColor,anchor=base east,inner sep=0pt, outer sep=0pt, scale=  0.80] at ( 25.08, 83.06) {40};

\node[text=drawColor,anchor=base east,inner sep=0pt, outer sep=0pt, scale=  0.80] at ( 25.08,112.75) {60};
\end{scope}
\begin{scope}
\path[clip] (  0.00,  0.00) rectangle (252.94,122.86);
\definecolor{drawColor}{gray}{0.20}

\path[draw=drawColor,line width= 0.6pt,line join=round] ( 27.28, 56.13) --
	( 30.03, 56.13);

\path[draw=drawColor,line width= 0.6pt,line join=round] ( 27.28, 85.82) --
	( 30.03, 85.82);

\path[draw=drawColor,line width= 0.6pt,line join=round] ( 27.28,115.51) --
	( 30.03,115.51);
\end{scope}
\begin{scope}
\path[clip] (  0.00,  0.00) rectangle (252.94,122.86);
\definecolor{drawColor}{RGB}{0,0,0}

\path[draw=drawColor,line width= 0.6pt,line join=round] ( 30.03, 29.10) --
	(247.44, 29.10);

\path[draw=drawColor,line width= 0.6pt,line join=round] (244.98, 27.67) --
	(247.44, 29.10) --
	(244.98, 30.52);
\end{scope}
\begin{scope}
\path[clip] (  0.00,  0.00) rectangle (252.94,122.86);
\definecolor{drawColor}{gray}{0.20}

\path[draw=drawColor,line width= 0.6pt,line join=round] ( 38.36, 26.35) --
	( 38.36, 29.10);

\path[draw=drawColor,line width= 0.6pt,line join=round] (116.17, 26.35) --
	(116.17, 29.10);

\path[draw=drawColor,line width= 0.6pt,line join=round] (193.99, 26.35) --
	(193.99, 29.10);
\end{scope}
\begin{scope}
\path[clip] (  0.00,  0.00) rectangle (252.94,122.86);
\definecolor{drawColor}{gray}{0.30}

\node[text=drawColor,anchor=base,inner sep=0pt, outer sep=0pt, scale=  0.80] at ( 38.36, 18.64) {0};

\node[text=drawColor,anchor=base,inner sep=0pt, outer sep=0pt, scale=  0.80] at (116.17, 18.64) {100};

\node[text=drawColor,anchor=base,inner sep=0pt, outer sep=0pt, scale=  0.80] at (193.99, 18.64) {200};
\end{scope}
\begin{scope}
\path[clip] (  0.00,  0.00) rectangle (252.94,122.86);
\definecolor{drawColor}{RGB}{0,0,0}

\node[text=drawColor,anchor=base,inner sep=0pt, outer sep=0pt, scale=  1.00] at (138.74,  7.44) {$d$};
\end{scope}
\begin{scope}
\path[clip] (  0.00,  0.00) rectangle (252.94,122.86);
\definecolor{drawColor}{RGB}{0,0,0}

\node[text=drawColor,rotate= 90.00,anchor=base,inner sep=0pt, outer sep=0pt, scale=  1.00] at ( 12.39, 73.23) {time (ms)};
\end{scope}
\begin{scope}
\path[clip] (  0.00,  0.00) rectangle (252.94,122.86);

\path[] ( 29.60, 88.58) rectangle (156.56,128.49);
\end{scope}
\begin{scope}
\path[clip] (  0.00,  0.00) rectangle (252.94,122.86);
\definecolor{drawColor}{RGB}{50,113,173}

\path[draw=drawColor,line width= 0.6pt,line join=round] ( 36.55,115.76) -- ( 48.11,115.76);
\end{scope}
\begin{scope}
\path[clip] (  0.00,  0.00) rectangle (252.94,122.86);
\definecolor{fillColor}{RGB}{50,113,173}

\path[fill=fillColor] ( 42.33,115.76) circle (  1.96);
\end{scope}
\begin{scope}
\path[clip] (  0.00,  0.00) rectangle (252.94,122.86);
\definecolor{drawColor}{RGB}{70,165,69}

\path[draw=drawColor,line width= 0.6pt,dash pattern=on 2pt off 2pt ,line join=round] ( 36.55,101.31) -- ( 48.11,101.31);
\end{scope}
\begin{scope}
\path[clip] (  0.00,  0.00) rectangle (252.94,122.86);
\definecolor{fillColor}{RGB}{70,165,69}

\path[fill=fillColor] ( 42.33,104.36) --
	( 44.97, 99.78) --
	( 39.69, 99.78) --
	cycle;
\end{scope}
\begin{scope}
\path[clip] (  0.00,  0.00) rectangle (252.94,122.86);
\definecolor{drawColor}{RGB}{0,0,0}

\node[text=drawColor,anchor=base west,inner sep=0pt, outer sep=0pt, scale=  0.80] at ( 55.06,113.00) {Lifted Variable Elimination};
\end{scope}
\begin{scope}
\path[clip] (  0.00,  0.00) rectangle (252.94,122.86);
\definecolor{drawColor}{RGB}{0,0,0}

\node[text=drawColor,anchor=base west,inner sep=0pt, outer sep=0pt, scale=  0.80] at ( 55.06, 98.55) {Variable Elimination};
\end{scope}
\end{tikzpicture}
	\caption{The average run times of \acl{ve} and \acl{lve}.}
	\label{fig:lifagu-plot-runtimes}
\end{figure}

In addition to the error measured by the \ac{kld}, we also report the run times of \acl{ve} on $G$ and \acl{lve} on the \ac{pfg} computed by \ac{lifg}, i.e., $G_{\textsc{\ac{lifg}}}$.
The average run times over all scenarios are shown in \cref{fig:lifagu-plot-runtimes}.
As expected, \acl{lve} is faster than \acl{ve} for larger graphs and the run time of \acl{lve} increases more slowly with increasing graph sizes than the run time of \acl{ve}.
Hence, \ac{lifg} not only allows to perform probabilistic inference at all, but also speeds up inference by allowing for lifting probabilistic inference.
Note that there are on average $17$ different groups of \acp{rv} over all settings with the largest group size being $205$ (for the setting of $d=256$), i.e., there are a lot of small groups (of size one) which diminish the advantage of \acl{lve} over \acl{ve}.
We could also obtain a more compact \ac{pfg} by merging groups that are not fully identical but similar to a given extent such that the resulting \ac{pfg} contains less different groups at the cost of a lower accuracy for query results.
Obtaining a more compact \ac{pfg} would most likely result in a higher speedup of \acl{lve} compared to \acl{ve}.

Finally, we remark that assuming there exists at least one group to which a new individual can be added is clearly a strong assumption that might not hold in practical settings.
We made this assumption for our experiments to guarantee a well-defined semantics of the model, as otherwise query answering would not be possible at all and hence, comparing \acp{kld} and run times could not be performed.
Despite this rather strong assumption, the experiments provide a first impression for the order of magnitude of the error induced by \ac{lifg}.
The results on the synthetic data used in this section are promising and suggest that \ac{lifg} performs well in practice.

\section{Conclusion} \label{sec:conclusion}
We introduce the \ac{lifg} algorithm to construct a lifted representation, denoted as a \ac{pfg}, for an \ac{fg} that possibly contains factors whose underlying potential mappings are unknown.
\Ac{lifg} is a generalisation of the \ac{acp} algorithm and allows to transfer potentials from known factors to unknown factors by identifying indistinguishable subgraph structures.
Under the assumption that for every unknown factor there exists at least one known factor such that they have an indistinguishable surrounding graph structure, \ac{lifg} is able to replace all unknown potential mappings in an \ac{fg} by known potential mappings.
To reduce ambiguity when grouping unknown factors with known factors, we introduce the concept of supporting background knowledge and show how it can be integrated into \ac{lifg}.

In future work, we aim to further generalise the \ac{acp} algorithm to allow for a small deviation between the potentials of two known factors $f_1$ and $f_2$ for $f_1$ and $f_2$ to be considered identical while at the same time maintaining a bounded error on probabilistic queries posed to the lifted model.

\section*{Acknowledgements}
This work is partially funded by the BMBF project AnoMed 16KISA057 and 16KISA050K.

\bibliographystyle{elsarticle-harv}
\bibliography{references.bib}
\end{document}